\documentclass[10pt,journal,compsoc]{IEEEtran}

\ifCLASSOPTIONcompsoc
\usepackage[nocompress]{cite}
\else
\usepackage{cite}
\fi

\ifCLASSINFOpdf
\else
\fi

\usepackage{epsfig}
\usepackage{graphicx}
\usepackage{amsmath}
\usepackage{amssymb}
\usepackage{enumitem}

\usepackage[dvipsnames,table]{xcolor}

\usepackage[pagebackref=true,breaklinks=true,letterpaper=true,colorlinks,bookmarks=false,urlcolor=red,citecolor=blue]{hyperref}
\usepackage{gensymb,graphics,subfigure,inputenc}
\usepackage[linesnumbered,algo2e,boxed]{algorithm2e}
\usepackage{ragged2e}
\graphicspath{{Figure/}}
\usepackage[figuresright]{rotating}
\usepackage{amsmath,amssymb,textcomp,subfigure,multirow,upgreek}
\usepackage{booktabs}
\usepackage{color,mdwlist}

\usepackage{makecell}
\usepackage{graphicx}
\usepackage{utfsym}
\usepackage{cleveref}
\usepackage{threeparttable,booktabs}
\usepackage{subcaption}
\usepackage{longtable}

\usepackage{wasysym}
\usepackage{utfsym}
\usepackage{amsbsy}
\usepackage{pifont}

\definecolor{DIMEBestText}{HTML}{D55E00}
\definecolor{DIMESecondText}{HTML}{0072B2}
\newcommand{\BestScore}[1]{\textcolor{DIMEBestText}{#1}}
\newcommand{\SecondScore}[1]{\textcolor{DIMESecondText}{#1}}

\newcommand{\fidup}[1]{\textcolor{gray}{\scriptsize\, $\uparrow$#1}}
\newcommand{\fiddown}[1]{\textcolor{gray}{\scriptsize\, $\downarrow$#1}}

\usepackage{threeparttablex}
\usepackage{array}

\definecolor{ManuscriptRed}{HTML}{8B0000}
\definecolor{ManuscriptBlack}{HTML}{000000}

\usepackage{graphicx,fontawesome}
\graphicspath{{Figures/}}

\usepackage{caption}
\usepackage{enumitem}
\setitemize{noitemsep,topsep=0pt,parsep=0pt,partopsep=0pt}

\makeatletter
\providecommand{\figcaption}{\def\@captype{figure}\caption}
\makeatother

\begin{document}

\title{DIME: Scaling Facial Representation Learning via Differential Masked Autoencoding}

\author{Hao~Yu, Haoyu~Chen, Hui~Wei, Yan~Jiang, Nicu~Sebe,
and~Guoying~Zhao,~\IEEEmembership{Fellow,~IEEE}
\IEEEcompsocitemizethanks{
\IEEEcompsocthanksitem H. Yu, H. Chen, H. Wei, Y. Jiang, and G. Zhao are with the ELLIS Institute Finland and CMVS, University of Oulu, Finland. N. Sebe is with DISI, University of Trento, Italy. Corresponding author: Guoying Zhao (e-mail: guoying.zhao@oulu.fi).
}
}

\markboth{IEEE Transactions on Pattern Analysis and Machine Intelligence,~Vol.~XX, No.~X, XXXXX~2026}%
{Yu \MakeLowercase{\textit{et al.}}: DIME: Scaling Facial Representation Learning via Differential Masked Autoencoding}

\IEEEtitleabstractindextext{%
\justifying

\begin{abstract}
Scaling up data and model capacity has driven substantial progress in visual representation learning. However, we observe that conventional pretext tasks exhibit diminishing returns on fine-grained facial analysis. We attribute this to a scaling bottleneck: these tasks model the appearance of each image and are dominated by identity, pose, and illumination, so the subtle within-identity facial differences that such analysis depends on contribute little to the training signal. We introduce DIME, a differential masked autoencoder for large-scale facial representation learning. DIME builds on a simple observation: two images of the same person share identity-specific appearance, so their difference exposes how the face changes. DIME therefore reconstructs both masked images in an identity-matched pair and, through an Explicit Differential Direction Supervision (EDDS) objective, aligns the direction of their reconstructed difference with the true difference, while source isolation makes each reconstruction depend only on its own image. The shared appearance cancels in this difference, and because EDDS ignores magnitude, subtle facial changes count as much as large pose or illumination changes. DIME uses identity labels only to form pretraining pairs, requires no dense annotations, and learns dense facial representations with emergent part-level semantics. We pretrain DIME on 20 million high-resolution facial images consolidated from existing face datasets. We show that DIME’s AU detection performance continues to improve as pretraining data scales from 6M to 20M images, with its BP4D F1 increasing by 5.9 points, while MAE gains only 0.5 points on the same subsets. DIME also outperforms specialized facial methods on several challenging face-related tasks: AU detection, AU intensity estimation, cross-domain anti-spoofing, landmark localization, head pose estimation, face parsing, and face generation. Our code is available on the project page: \url{https://cv-ac.github.io/DIME_page}.
\end{abstract}

}

\IEEEaftertitletext{%
  \par\vspace*{-2.8\baselineskip}
  \noindent
  \begin{minipage}{\textwidth}
    \normalfont\normalsize
    \centering
    \setlength{\abovecaptionskip}{0pt}
    \setlength{\belowcaptionskip}{0pt}

    \includegraphics[width=\linewidth]{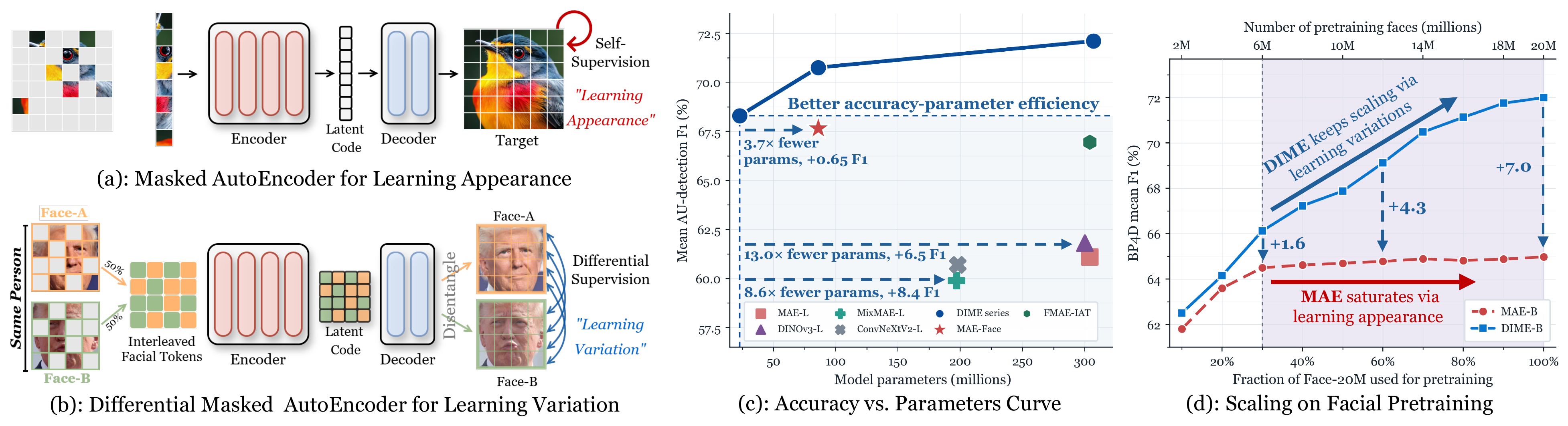}
    \par
    \setcounter{figure}{0}
    \figcaption{\textbf{Scaling facial pretraining via DIME.}
    \textit{(a)} MAE reconstructs each image in isolation to learn appearance.
    \textit{(b)} DIME treats two observations of the \emph{same} person
    as a single learning unit and explicitly supervises their difference
    to model how a face changes.
    \textit{(c)} Compared with generic visual encoders, DIME is far more efficient in facial representation learning tasks such as AU detection
    (mean performance over BP4D \cite{zhang2014bp4d} and BP4D+ \cite{zhang2016multimodal}).
    \textit{(d)} Pretrained with the same settings and the same amount of facial data,
    MAE \cite{he2022masked} saturates at 6M faces, whereas DIME keeps scaling
    all the way to 20M.}
    \label{fig:1}
  \end{minipage}
  \par\vspace{0.5\baselineskip}
}

\maketitle
\IEEEdisplaynontitleabstractindextext
\IEEEpeerreviewmaketitle

\setcounter{figure}{1}

\section{Introduction}
\label{sec:introduction}

\IEEEPARstart{F}{aces} support a broad range of applications beyond identity recognition~\cite{wright2008robust,kim202650}. Through action unit (AU) analysis~\cite{ma2024facial}, expression understanding~\cite{zhao2007dynamic}, face editing~\cite{melnik2024face}, and facial attack detection~\cite{yu2025learning}, they now shape fields ranging from healthcare and human-computer interaction to immersive media and biometric security. Despite the different prediction targets, these tasks share a perceptual basis of fine-grained deformation, texture, and subtle chromatic variation \cite{zhao2007dynamic,zhao2011facial,wang2025fsfm} in faces, which demand representation learning beyond the macro-level identity. Annotating these task-specific cues is costly and often requires domain expertise, leaving only limited labeled data for training task-specific models that are prone to identity and dataset shortcuts. This motivates a generic facial foundation model that preserves both global facial structure and subtle local variations, providing reusable representations of expression, configuration, and surface appearance across subjects.

\IEEEpubidadjcol

\begin{figure*}[!t]
  \centering
  \setlength{\abovecaptionskip}{-1pt}
  \includegraphics[width=\textwidth]{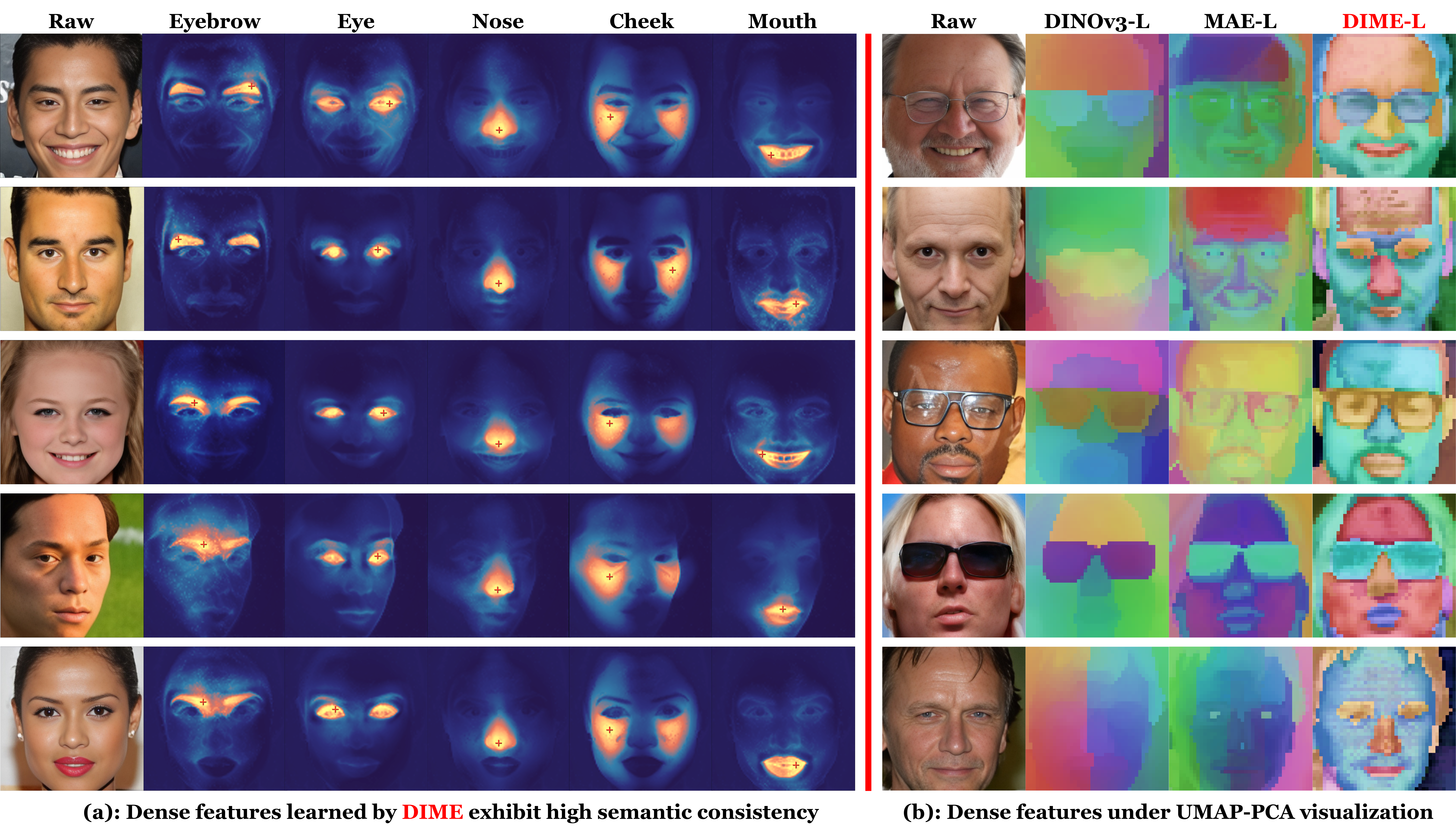}
\caption{\textbf{Emergent capacity of DIME.}
\textit{(a)} Cosine similarity between a marked query token (red cross) and all patches, demonstrating that DIME learns robust part-level semantic consistency across distinct facial parts (brighter colors indicate higher similarity).
\textit{(b)} UMAP-PCA \cite{mcinnes2018umap} visualizations of dense features from DIME-L, DINOv3-L~\cite{simeoni2025dinov3}, and MAE-L~\cite{he2022masked}. DIME exhibits more semantically differentiated and structurally aligned facial representations.}
  \label{fig:DIME_semantic}
  \vspace{-1.8em}
\end{figure*}

Self-supervised learning~\cite{he2022masked,simeoni2025dinov3,liu2023mixmae} offers a scalable route to such a model, drawing on massive face images without costly dense annotations. Self-distillation methods~\cite{caron2021emerging,oquab2023dinov2,simeoni2025dinov3} align predictions across augmented views, and masked image modeling~\cite{he2022masked,xie2022simmim,liu2023mixmae,wang2025bootstrap} predicts missing content from visible context. Both work well on natural images, whose transferable structure is organized around objects, spatial layouts, and scene semantics.

Applied to faces, however, scaling existing self-supervised paradigms encounters a bottleneck for fine-grained facial representation learning. We measure it on AU detection: AUs~\cite{ekman1976measuring} encode the activations of individual facial muscles, and their combinations describe nearly all visible facial dynamics. Detecting AUs requires the subtle and identity-independent cues that many fine-grained facial tasks build on, so AU detection performance serves as a proxy for the quality of learned fine-grained facial representations. Specifically, as shown in Fig.~\ref{fig:1}(d), we pretrain the masked autoencoder (MAE)~\cite{he2022masked} on progressively larger sets of face images and fine-tune each pretrained encoder for AU detection on BP4D~\cite{zhang2014bp4d} under an identical protocol. Its F1 score begins to saturate at 6M pretraining images and improves by less than 1\% as the dataset more than triples to 20M. This plateau suggests that scaling the pretraining mostly broadens the coverage of coarse appearance without better capturing these fine-grained cues of facial dynamics.

We trace this plateau to the way pixel reconstruction weights the factors of facial variation. Conventional masked image modeling regresses the normalized pixels of each masked patch with a squared loss, so the optimal prediction is their conditional mean given the visible patches. Modeling a factor then lowers the loss by the share of patch variance that it explains and that the visible patches can predict. A model with finite capacity and training time therefore favors factors that explain a large share of each patch, as the optimal low-rank predictor does in the linear case~\cite{izenman1975reduced}. On aligned faces, most of this variance comes from the shared facial layout~\cite{blanz1999morphable}, identity-specific appearance, and large pose and illumination changes; between face images of the same person, such changes can even exceed the differences between different people~\cite{adini1997face}. The subtle changes that fine-grained tasks rely on, such as the local deformations of AUs, explain only a small share of each patch and contribute little to the loss. More images thus mainly refine the prediction of the large factors and leave the subtle ones underweighted, consistent with the plateau in Fig.~\ref{fig:1}(d). Contrastive and self-distillation objectives face a different limit. By making the features of two augmented views agree, they discard whatever the augmentations change~\cite{xiao2021should}, including color and geometric cues that facial tasks may need; contrastive facial features, for example, tend to lose pose detail~\cite{liu2023pose}.

To scale fine-grained facial representation learning, we propose \textbf{DIME}, a foundation model trained with \textbf{Di}fferential \textbf{M}asked Auto\textbf{e}ncoding (Fig.~\ref{fig:1}(b)). Unlike conventional masked image modeling, which learns the appearance of each image, DIME learns by reproducing how two face images of the same person differ. In their difference, the common-mode component cancels, including the facial layout and identity-specific appearance that dominate pixel reconstruction. What remains is the variation within a single identity, which mixes large pose changes with subtle facial deformations, and we turn it into an explicit learning target.

DIME realizes this idea through three designs. First, to keep the change from being copied, we mask both images and reconstruct them with the same network, so the change at hidden patches must be inferred from visible context. The two reconstructions are compared only through their difference, so their shared appearance never enters the comparison. Second, to keep subtle changes from being outweighed, we supervise this difference with Explicit Differential Direction Supervision (EDDS), which requires the reconstructed difference to point in the same direction as the true difference between the source images. Because EDDS disregards magnitude, missing a subtle change costs as much as missing a pronounced one, whereas a squared loss discounts each change by its energy. Third, we use source isolation masks that restrict attention to tokens of the same image, so neither image can supply the other's hidden patches. With these shortcuts closed, the model can lower EDDS mainly by inferring how a masked patch has changed from how the visible patches of the two images have changed. Pretraining is thereby steered toward fine-grained facial variations regardless of personal appearance.

Our experiments show that DIME keeps improving AU detection with more pretraining data and, using only the cheap identity labels, learns dense and fine-grained facial representations. Under the same training settings, DIME-B gains 5.9 F1 points as the pretraining data grows from 6M to 20M images, whereas MAE-B gains only 0.5 (Fig.~\ref{fig:1}(d)). This gap matches the analysis above: more data keeps adding fine-grained facial variation for DIME to learn, whereas for MAE it mainly refines the appearance factors. Even the smallest DIME-S surpasses general backbones~\cite{simeoni2025dinov3,he2022masked,liu2023mixmae,woo2023convnext} and facial foundation models~\cite{ning2024representation,ma2024facial} in AU detection with $3.7\times$ to $13\times$ fewer parameters (Fig.~\ref{fig:1}(c)). Following the visualization protocol of DINOv3~\cite{simeoni2025dinov3}, we show \textit{emergent} part-level semantics in DIME: its patch features stay within semantically consistent facial parts (Fig.~\ref{fig:DIME_semantic}(a)) and align structurally across subjects (Fig.~\ref{fig:DIME_semantic}(b)). Overall, DIME reaches state-of-the-art performance on seven face-related tasks: AU detection, AU intensity estimation, cross-domain anti-spoofing, landmark localization, head pose estimation, face parsing, and face generation. Our contributions are:

\begin{itemize}
\item We identify a scaling bottleneck in facial pretraining and propose DIME, which makes facial data scaling effective for learning fine-grained representations.

\item We introduce EDDS to turn subtle facial variations into learning targets, enabling DIME to learn fine-grained facial semantics from large-scale face data with only cheap identity labels and no dense annotations.

\item We show that the benefits of scaling DIME's pretraining transfer effectively across a broad suite of facial analysis and generation tasks, where DIME outperforms generic visual foundation models, existing facial foundation models, and domain-specialized methods.
\end{itemize}

\section{Related Work}
\label{sec:related_work}

\subsection{Self-Supervised Visual Pretraining}
\label{sec:related_ssl}

Self-supervised visual pretraining has two main paradigms: enforcing consistency across views and predicting missing visual content. Contrastive and self-distillation frameworks~\cite{chen2020simple,caron2021emerging} learn representations by maximizing agreement between augmented views. DINOv2~\cite{oquab2023dinov2} adds masked patch prediction to self-distillation to improve transfer, and DINOv3~\cite{simeoni2025dinov3} introduces Gram anchoring to prevent the degradation of dense features during long training.

Masked image modeling learns representations by predicting missing visual content from visible context. MAE~\cite{he2022masked} introduces an efficient asymmetric architecture whose encoder processes only visible tokens. Later work predicts other targets such as local descriptors~\cite{wei2022masked}, simplifies the prediction scheme so that it also applies to hierarchical backbones~\cite{xie2022simmim}, or, as in I-JEPA~\cite{assran2023self}, predicts the latent representations of target regions in place of their pixels. For masked image modeling on natural images, gains from more data depend mainly on model size and training length~\cite{xie2023data} under a fixed objective. Several methods pair two images in masked modeling. MixMAE~\cite{liu2023mixmae} mixes tokens from two randomly sampled images and restricts attention to tokens of the same image. Its objective is a sum of per-image reconstruction losses, so the pairing mainly replaces encoder-side mask tokens and improves pretraining efficiency. Cross-view completion instead feeds the second image into the reconstruction: CroCo~\cite{weinzaepfel2022croco} completes a masked view from its visible patches and a second, unmasked view of the same scene, and SiamMAE~\cite{gupta2023siamese} predicts the masked patches of a future video frame from an unmasked past frame through decoder cross-attention. DIME pairs two observations of the same identity, keeps them apart in the forward pass, and couples them only through EDDS, a loss on the direction of the difference between the two reconstructions.

\subsection{Facial Representation Learning and Pretraining}
\label{sec:related_face}

Domain-specific facial pretraining adapts general self-supervised objectives with facial priors and specialized architectures. FaRL~\cite{zheng2022general} combines image-text contrastive learning with masked image modeling, MAE-Face~\cite{ma2024facial} uses masked image modeling for action unit analysis, FSFM~\cite{wang2025fsfm} integrates region-specific masking with self-distillation for biometric security, and PaCo-FR~\cite{xie2026paco} learns a patch-pixel aligned codebook end to end during pretraining. Other single-image strategies explore privacy-preserving reconstruction~\cite{wang2022facemae} or landmark-guided local tokenization~\cite{sun2024lafs}.

Video-based frameworks exploit temporal continuity across observations: MARLIN~\cite{cai2023marlin} performs spatiotemporal reconstruction, MAE-DFER~\cite{sun2023mae} predicts inter-frame differences, and AU-vMAE~\cite{jin2024vmae} uses video pretraining with AU-guided fine-tuning. Earlier methods learn facial change from video frame pairs: FAb-Net~\cite{wiles2018self} maps a source frame to a target frame of the same face through a predicted flow field, and TCAE~\cite{li2019self} separates AU-related from head-motion-related movements while transforming one face into another. Both need two frames of one person from the same video. Another line of work disentangles the factors of facial variation. PCL~\cite{liu2023pose} separates pose-dependent from pose-invariant features, LatentFace~\cite{he2023generative} isolates identity from expression with 3D-aware autoencoding and temporally averaged identity latents, and FMAE-IAT~\cite{ning2024representation} applies identity-adversarial regularization during downstream fine-tuning. Other methods add language supervision: FaceLLM~\cite{shahreza2025facellm} adapts a multimodal language model with generated facial question-answer pairs, and GPTFace~\cite{li2025gptface} combines masked image and language modeling with image-text matching for facial understanding and generation.

Unlike these methods, DIME learns from still images grouped by identity and needs no latent factorization, adversarial loss, video, or text. During pretraining, DIME reconstructs token-mixed image pairs and, through EDDS, aligns the direction of each reconstructed patch difference with the true difference. DIME thus turns within-identity variation into an explicit pretraining target, while its encoder needs only a single image for downstream transfer.

\begin{figure*}[!t]
\centering
  \setlength{\abovecaptionskip}{0pt}
\includegraphics[width=1.0\textwidth]{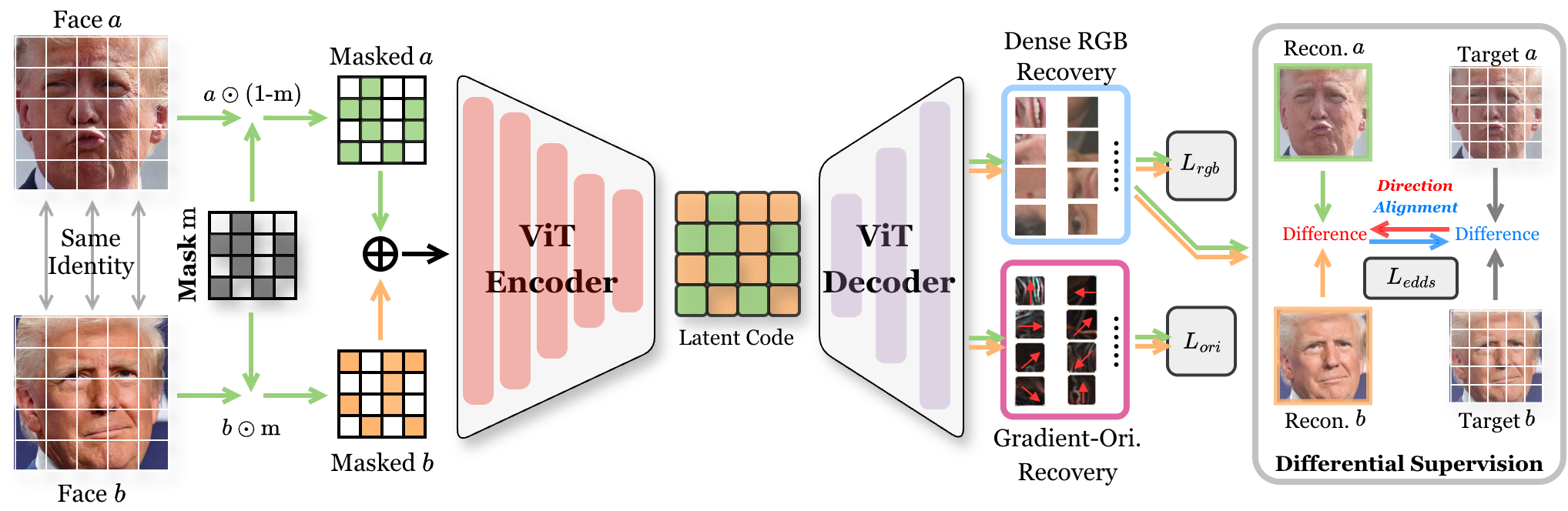}
\caption{\textbf{Overall architecture of DIME.} For simplicity, we illustrate one mixed token grid. In practice, the two reciprocal mixed grids and their complementary decoder branches are batched and processed in parallel to improve efficiency.}
\label{fig:framework}
\vspace{-1.5em}
\end{figure*}

\section{Proposed Method}

\subsection{Problem Formulation}
\label{sec:dime_formulation}

DIME learns facial representations from pairs of images of the same person. Under raw pixel reconstruction, predictable facial geometry and appearance account for most of the loss, and the changes that matter for tasks such as AU analysis appear as small, localized deviations. DIME therefore adds a loss on the difference between the two reconstructions, so that these deviations receive a training signal of their own. To train DIME, we first organize the training data into a set of same-identity image pairs,
\begin{equation}
\mathcal{P}=\bigl\{(\mathbf{x}^{a}_{n},\mathbf{x}^{b}_{n})\bigr\}_{n=1}^{N},
\qquad
\operatorname{id}(\mathbf{x}^{a}_{n})=\operatorname{id}(\mathbf{x}^{b}_{n}),
\label{eq:dime_pair_set}
\end{equation}
where $\mathbf{x}^{a}_{n},\mathbf{x}^{b}_{n}\in\mathbb{R}^{3\times H\times W}$ are distinct observations of one person, $n$ indexes pairs, and $\operatorname{id}$ denotes the identity label.

\noindent\textbf{Supervision used by DIME.} We use identity labels only to form pairs; the network never receives or predicts the identity. The training signal specific to DIME is the difference between the two reconstructions, in which the appearance shared by the identity cancels (Sec.~\ref{sec:dime_formal}). What DIME adds to masked reconstruction therefore comes from how the two faces differ in expression and pose under shared identity.

The backbone of DIME comprises a patch tokenizer $\mathcal{E}_{p}$, a ViT encoder $\mathcal{F}$, and a lightweight decoder $\mathcal{G}$~\cite{he2022masked}. Each image yields $L$ patches of $P$ pixels, with $LP=HW$, and tokens $\mathbf{T}^{i}=\mathcal{E}_{p}(\mathbf{x}^{i})\in\mathbb{R}^{L\times C}$ for $i\in\{a,b\}$. Let $\mathbf{Y}^{i}\in\mathbb{R}^{L\times 3P}$ stack the raw target patches of $\mathbf{x}^{i}$, with rows $(\mathbf{y}^{i}_{\ell})^{\top}$, and let $\hat{\mathbf{Y}}^{i}$ stack the corresponding predictions, with rows $(\hat{\mathbf{y}}^{i}_{\ell})^{\top}$, expressed in per-patch normalized coordinates (Sec.~\ref{sec:dime_losses}). Pretraining minimizes the loss $\mathcal{L}_{\mathrm{DIME}}$ of Eq.~\eqref{eq:dime_total_objective}, averaged over same-identity pairs and random source masks.

Fig.~\ref{fig:framework} summarizes the pipeline. Sec.~\ref{sec:dime_source_recovery} reconstructs both images from isolated masked views, Sec.~\ref{sec:dime_losses} defines the losses on the recovered pair, and Sec.~\ref{sec:dime_formal} analyzes the properties of EDDS. Below, we omit the pair index; tildes denote mixed token grids, and hats denote predictions.

\subsection{Encoding and Reciprocal Decoding}
\label{sec:dime_source_recovery}

Unlike the single-image learning unit of conventional masked autoencoding~\cite{he2022masked,xie2022simmim}, DIME places two same-identity observations in mixed token grids. A random binary mask $\mathbf{m}\in\{0,1\}^{L}$, with a fraction $r$ of ones ($r=0.5$ by default), selects the source at each spatial position. For $i,j\in\{a,b\}$ with $i\neq j$, the source-mixing operation is
\begin{equation}
\operatorname{mix}(\mathbf{T}^{i},\mathbf{T}^{j};\mathbf{m})
=
(\mathbf{1}_{L}-\mathbf{m})\odot\mathbf{T}^{i}
+
\mathbf{m}\odot\mathbf{T}^{j},
\label{eq:dime_reciprocal_mix}
\end{equation}
where $\mathbf{1}_{L}$ is the length-$L$ all-ones vector, and $\odot$ is elementwise multiplication with the mask broadcast over channels. Positions with $m_p=0$ contain the first source; those with $m_p=1$ contain the second. Recovering both provides the paired predictions for differential supervision.

Following MixMAE~\cite{liu2023mixmae}, mixing lets one encoder pass process the visible parts of two images without mask tokens. For each pair, DIME applies the mixture reciprocally in both source orders under the same partition:
\begin{equation}
\begin{aligned}
\widetilde{\mathbf{T}}^{a}
&=
\operatorname{mix}(\mathbf{T}^{a},\mathbf{T}^{b};\mathbf{m}),\\
\widetilde{\mathbf{T}}^{b}
&=
\operatorname{mix}(\mathbf{T}^{b},\mathbf{T}^{a};\mathbf{m}),
\end{aligned}
\label{eq:dime_reciprocal_realization}
\end{equation}
and both grids pass through the shared encoder:
\begin{equation}
\mathbf{Z}^{i}
=
\mathcal{F}(\widetilde{\mathbf{T}}^{i};\mathbf{m}),
\qquad
i\in\{a,b\},
\label{eq:dime_shared_encoder}
\end{equation}
where $\mathbf{Z}^{i}\in\mathbb{R}^{L\times C}$. Across the two grids, each image is visible at complementary positions, so every patch of each image can be predicted from a grid in which it is hidden.

To prevent one observation from supplying the other's missing content, we follow the masked self-attention of MixMAE~\cite{liu2023mixmae}, in which every encoder block adds the following bias to the attention logits between positions $p$ and $q$:
\begin{equation}
A_{p,q}=
\begin{cases}
0, & m_{p}=m_{q},\\[2pt]
-\infty, & m_{p}\neq m_{q}.
\end{cases}
\label{eq:dime_source_attention_bias}
\end{equation}

With this bias, attention connects only tokens from the same source; we call this source isolation. The remaining encoder operations are tokenwise, and no global token is shared, so each encoded token depends on its own image alone. Without isolation, each face could supply the evidence masked in its similar partner, and the change could be copied instead of inferred.

For each grid $i\in\{a,b\}$, the shared decoder $\mathcal{G}$ runs on two complementary inputs: branch~$k\in\{0,1\}$ keeps the projected encoder features at positions with $m_p=k$ and fills the remaining positions with a learned mask token. Each decoder input therefore contains features from only one source. Let $\mathbf{R}^{i}_{k}\in\mathbb{R}^{L\times 3P}$ denote the normalized output of branch~$k$. We assemble the two source predictions as
\begin{equation}
\begin{aligned}
\hat{\mathbf{Y}}^{a}
&=
\mathbf{m}\odot\mathbf{R}^{a}_{0}
+
(\mathbf{1}_{L}-\mathbf{m})\odot\mathbf{R}^{b}_{1},\\
\hat{\mathbf{Y}}^{b}
&=
\mathbf{m}\odot\mathbf{R}^{b}_{0}
+
(\mathbf{1}_{L}-\mathbf{m})\odot\mathbf{R}^{a}_{1}.
\end{aligned}
\label{eq:dime_complementary_unmixing}
\end{equation}

For $a$, positions with $m_p=1$ are predicted from its visible context at $m_p=0$, and vice versa; source $b$ is recovered symmetrically. Each assembled patch comes from a branch in which its corresponding position is masked.

\subsection{Differential Supervision}
\label{sec:dime_losses}

DIME combines three losses on the recovered pair: EDDS on the difference between the two reconstructions, a gradient-orientation loss on local structure, and the normalized-pixel loss of MAE on each reconstruction.

\noindent\textbf{Difference-direction supervision.} Reconstructing each image separately can yield two plausible faces whose difference is wrong. When the change between them is small, such as a slight raising of the mouth corners, small errors in the two reconstructions can erase or even reverse it at little cost to the reconstruction loss. EDDS addresses this issue by requiring the reconstructed pair to change in the same local direction as the original pair. Specifically, for source $i\in\{a,b\}$ and patch $\ell$, let $\mathbf{y}^{i}_{\ell}\in\mathbb{R}^{3P}$ denote the raw target pixels and $\hat{\mathbf{y}}^{i}_{\ell}$ the corresponding prediction in normalized patch coordinates. Let $\mu^{i}_{\ell}$ and $\nu^{i}_{\ell}$ be the mean and variance over all $3P$ target values, and let $\varepsilon_{\mathrm{n}}>0$ be a numerical stabilizer. Because the two sources are normalized using different patch statistics, their predictions must be expressed in a common raw-pixel space before differencing:
\begin{equation}
\hat{\mathbf{y}}^{i,\mathrm{raw}}_{\ell}
=
\sqrt{\nu^{i}_{\ell}+\varepsilon_{\mathrm{n}}}\,
\hat{\mathbf{y}}^{i}_{\ell}
+
\mu^{i}_{\ell}\mathbf{1}_{3P},
\label{eq:dime_inverse_normalization}
\end{equation}
where the target statistics are computed from the target patch, carry no gradient, and are not provided to the encoder. The target and recovered differences are
\begin{equation}
\begin{aligned}
\boldsymbol{\Delta}_{\ell}
&=
\operatorname{reshape}_{P\times 3}
\bigl(\mathbf{y}^{a}_{\ell}-\mathbf{y}^{b}_{\ell}\bigr),\\
\hat{\boldsymbol{\Delta}}_{\ell}
&=
\operatorname{reshape}_{P\times 3}
\bigl(\hat{\mathbf{y}}^{a,\mathrm{raw}}_{\ell}
-\hat{\mathbf{y}}^{b,\mathrm{raw}}_{\ell}\bigr).
\end{aligned}
\label{eq:dime_pairwise_difference}
\end{equation}
Both terms of $\hat{\boldsymbol{\Delta}}_{\ell}$ are recovered patches. Raw differences may contain spatially uniform per-channel offsets that carry little useful information about local facial change. We thus remove this redundant component using the centering matrix $\mathbf{H}_{P}=\mathbf{I}_{P} -\tfrac{1}{P}\mathbf{1}_{P}\mathbf{1}_{P}^{\top}$, which satisfies $\mathbf{H}_{P}(\boldsymbol{\Delta}_{\ell}+\mathbf{1}_{P}\boldsymbol{\beta}^{\top})=\mathbf{H}_{P}\boldsymbol{\Delta}_{\ell}$ for any $\boldsymbol{\beta}\in\mathbb{R}^{3}$. Centering therefore removes constant RGB offsets within each patch and keeps the spatial pattern of change. It does not imply invariance to general illumination variation. We define the centered target and predicted residuals as
$\boldsymbol{\delta}_{\ell}
=\mathbf{H}_{P}\boldsymbol{\Delta}_{\ell}$ and
$\hat{\boldsymbol{\delta}}_{\ell}
=\mathbf{H}_{P}\hat{\boldsymbol{\Delta}}_{\ell}$. With the Frobenius inner product $\langle\cdot,\cdot\rangle_F$ and norm $\|\cdot\|_F$, EDDS minimizes their directional discrepancy:
\begin{equation}
\mathcal{L}_{\mathrm{edds}}
=
\frac{1}{\max(1,|\Gamma|)}
\sum_{\ell\in\Gamma}
\left[
1-
\frac{
\left\langle
\hat{\boldsymbol{\delta}}_{\ell},
\boldsymbol{\delta}_{\ell}
\right\rangle_F
}{
\left\|\hat{\boldsymbol{\delta}}_{\ell}\right\|_F
\left\|\boldsymbol{\delta}_{\ell}\right\|_F
+\varepsilon_{\mathrm{e}}
}
\right].
\label{eq:dime_edds_loss}
\end{equation}

Because $\mathbf{H}_{P}$ removes per-channel constants, the means $\mu^{i}_{\ell}$ cancel in $\hat{\boldsymbol{\delta}}_{\ell}$, and only the scales $\sqrt{\nu^{i}_{\ell}+\varepsilon_{\mathrm{n}}}$ enter EDDS. As in the normalized-pixel loss of MAE, these scales come from the targets, so EDDS compares the predicted normalized patterns after rescaling them by the true patch contrasts. EDDS therefore does not supervise the change of patch contrast between the two images; that part of the difference is supplied by the target statistics.

Here $\varepsilon_{\mathrm{e}}>0$ is a small stabilizer; with $\varepsilon_{\mathrm{e}}=0$, each bracketed term is invariant to any positive rescaling of $\hat{\boldsymbol{\delta}}_{\ell}$ or $\boldsymbol{\delta}_{\ell}$, so EDDS compares only the directions of the predicted and target changes. The support $\Gamma$ is selected from the target residuals alone and therefore does not depend on the model prediction. Let $d_{\ell}=\|\boldsymbol{\delta}_{\ell}\|_F$, $\bar d=L^{-1}\sum_{\ell}d_{\ell}$, and $\bar d_{\mathcal{B}}$ be the mean of $\bar d$ over the pairs in the minibatch. For $\rho,\kappa\in(0,1)$, $\Gamma$ keeps the top $(1-\rho)$ fraction of patches by $d_{\ell}$ when $\bar d>\kappa\bar d_{\mathcal{B}}$, and is empty otherwise. This gate skips pairs that change little relative to the minibatch, such as near-duplicate images, and within each remaining pair drops the patches that change least, where the direction of change is least reliable. Within $\Gamma$, every patch counts equally regardless of how much it changes, and empty supports contribute zero. Finally, EDDS is invariant to pair order: swapping $a$ and $b$ negates both $\boldsymbol{\delta}_{\ell}$ and $\hat{\boldsymbol{\delta}}_{\ell}$ and leaves the loss and its support unchanged, so it needs neither a temporal order nor a reference expression.

\noindent\textbf{Gradient-orientation recovery.}
While EDDS captures variation across observations, fine-grained facial evidence also depends on local oriented structure. We therefore add a decoder head that predicts horizontal and vertical gradients per color channel. Its outputs are assembled from the decoder branches as in Eq.~\eqref{eq:dime_complementary_unmixing}, so every predicted gradient also comes from a branch in which its position was masked.

Specifically, a fixed Sobel operator $\mathcal{S}$ computes the target gradients $\mathbf{G}^{i}=\mathcal{S}(\mathbf{x}^{i})$ for $i\in\{a,b\}$. At pixel $u$ and channel $c$, the gradient $\mathbf{G}^{i}_{u,c}\in\mathbb{R}^{2}$ has magnitude $g^{i}_{u,c}=\|\mathbf{G}^{i}_{u,c}\|_{2}$ and stabilized direction $\mathbf{v}^{i}_{u,c}=\mathbf{G}^{i}_{u,c}/(g^{i}_{u,c}+\varepsilon_{\mathrm{g}})$ with a small $\varepsilon_{\mathrm{g}}>0$; the predicted direction $\hat{\mathbf{v}}^{i}_{u,c}$ is defined likewise from the predicted gradient $\hat{\mathbf{G}}^{i}_{u,c}$. Because gradient direction is unreliable in nearly flat regions, we restrict supervision to
$\Omega=\{(i,u,c):i\in\{a,b\},\,
g^{i}_{u,c}>\tau_{i,c}\}$,
where $\tau_{i,c}$ is the $\rho$-quantile of target gradient magnitudes in channel $c$ of image $i$, with the same $\rho$ as in EDDS. The orientation objective is
\begin{equation}
\mathcal{L}_{\mathrm{ori}}
=
\frac{1}{\max(1,|\Omega|)}
\sum_{(i,u,c)\in\Omega}
\left(
1-
(\hat{\mathbf{v}}^{i}_{u,c})^{\top}
\mathbf{v}^{i}_{u,c}
\right).
\label{eq:dime_orientation_loss}
\end{equation}

Like EDDS, $\mathcal{L}_{\mathrm{ori}}$ scores direction only: target magnitude selects the supervised locations but does not weight their errors, so strong edges do not dominate the loss through magnitude alone. An empty support contributes zero.

\noindent\textbf{RGB reconstruction.}
We retain the normalized-pixel reconstruction objective of MAE~\cite{he2022masked} as a global reconstruction anchor. Specifically, each raw target patch is standardized as \scalebox{0.90}{$\mathcal{N}(\mathbf{y}^{i}_{\ell}) = (\mathbf{y}^{i}_{\ell} - \mu^{i}_{\ell}\mathbf{1}_{3P})/\sqrt{\nu^{i}_{\ell} + \varepsilon_{\mathrm{n}}}$}, and $\mathcal{L}_{\mathrm{rgb}}$ is the mean squared error between $\hat{\mathbf{y}}^{i}_{\ell}$ and $\mathcal{N}(\mathbf{y}^{i}_{\ell})$, averaged over both sources, all $L$ patches, and their $3P$ pixel values.

\noindent\textbf{Full objective.} The three objectives are combined as
\begin{equation}
\mathcal{L}_{\mathrm{DIME}}
=
\lambda_{\mathrm{edds}}\,\mathcal{L}_{\mathrm{edds}}
+
\lambda_{\mathrm{ori}}\,\mathcal{L}_{\mathrm{ori}}
+
\mathcal{L}_{\mathrm{rgb}},
\label{eq:dime_total_objective}
\end{equation}
where $\lambda_{\mathrm{edds}}$ and $\lambda_{\mathrm{ori}}$ are nonnegative balance factors.

\subsection{Properties and Discussion of EDDS}
\label{sec:dime_formal}
\begin{figure}[!t]
  \centering
  \setlength{\abovecaptionskip}{4pt}
  \includegraphics[width=\linewidth]{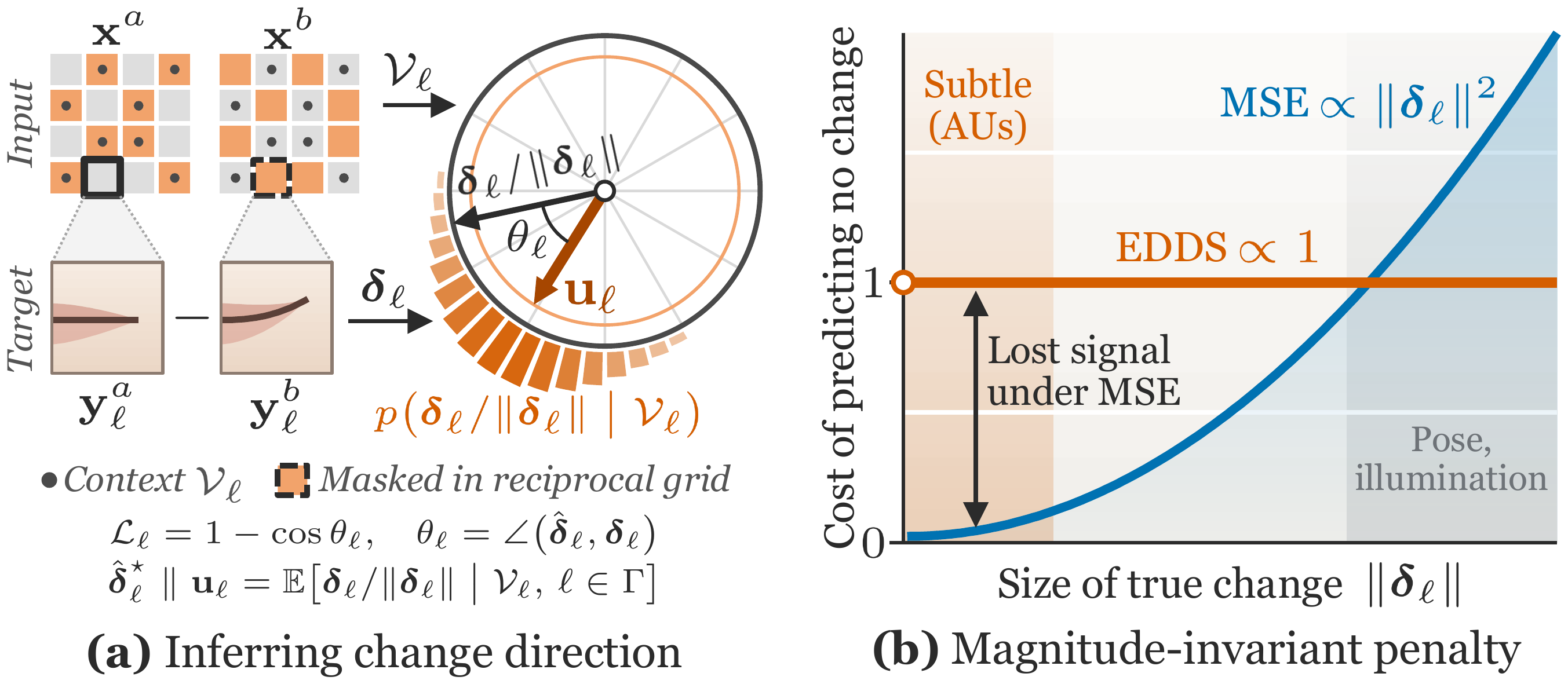}
\caption{\textbf{EDDS in masked image modeling.} (a) From the context $\mathcal{V}_{\ell}$ (dots) of two complementary masked views, the model infers the direction of change $\boldsymbol{\delta}_{\ell}$ at the hidden patch $\ell$. EDDS scores a prediction by $1-\cos\theta_{\ell}$, so the optimum points along the posterior mean $\mathbf{u}_{\ell}$, whose length reflects predictability. (b) Under MSE, predicting no change costs $\propto\|\boldsymbol{\delta}_{\ell}\|^{2}$, which nearly vanishes for subtle changes such as AUs; under EDDS it costs 1 at every supported patch, whatever the magnitude, recovering the lost signal.}
  \label{fig:edds_concept}
  \vspace{-1.8em}
\end{figure}

We now examine what EDDS asks the encoder to learn through two properties: it receives no signal from the appearance shared by a pair, and its cost of missing a change does not depend on how large the change is. Fig.~\ref{fig:edds_concept} illustrates how these properties play out when the change must be inferred from masked context, which also decides how EDDS treats noise and nuisance change.

\noindent\textbf{Shared appearance receives no EDDS gradient.}
Let $\hat{\mathbf{Q}}^{i}_{\ell}\in\mathbb{R}^{P\times3}$ be the recovered raw patch of image $i$, so that $\hat{\boldsymbol{\delta}}_{\ell}=\mathbf{H}_{P}(\hat{\mathbf{Q}}^{a}_{\ell}-\hat{\mathbf{Q}}^{b}_{\ell})$, let $\mathcal{L}_{\mathrm{edds},\ell}$ be the $\ell$-th term of Eq.~\eqref{eq:dime_edds_loss}, and let $\mathbf{q}_{\ell}=\nabla_{\hat{\boldsymbol{\delta}}_{\ell}}\mathcal{L}_{\mathrm{edds},\ell}$. Since $\Gamma$ depends only on the targets,
\begin{equation}
\nabla_{\hat{\mathbf{Q}}^{a}_{\ell}}\mathcal{L}_{\mathrm{edds},\ell}
=\mathbf{H}_{P}\mathbf{q}_{\ell}
=-\nabla_{\hat{\mathbf{Q}}^{b}_{\ell}}\mathcal{L}_{\mathrm{edds},\ell}.
\label{eq:dime_relational_gradient}
\end{equation}
The two gradients are equal and opposite, so they cancel along the common mode $\hat{\mathbf{Q}}^{a}_{\ell}+\hat{\mathbf{Q}}^{b}_{\ell}$: adding any pattern to both reconstructions leaves $\mathcal{L}_{\mathrm{edds}}$ unchanged. Shared appearance, which dominates pixel reconstruction, therefore cannot lower EDDS; only the difference between the two reconstructions can. Each gradient also depends on both images through $\mathbf{q}_{\ell}$, so EDDS updates the prediction of one image according to how it should differ from the other. The RGB loss $\mathcal{L}_{\mathrm{rgb}}$ anchors the shared appearance, and the two losses thus divide the work: reconstruction anchors what a face looks like, and EDDS supervises how it changes.

\noindent\textbf{Small changes are not discounted.}
EDDS depends only on the direction of the difference. Predicting no change ($\hat{\boldsymbol{\delta}}_{\ell}=\mathbf{0}$) therefore costs one at every supported patch, however small the true change is, whereas a squared loss on the difference would cost $\|\boldsymbol{\delta}_{\ell}\|_F^{2}$ and, like $\mathcal{L}_{\mathrm{rgb}}$, weight patches by the energy of their change (Fig.~\ref{fig:edds_concept}(b)). For subtle changes such as AU activations, a squared loss thus offers almost no incentive to predict the change, whereas EDDS offers the same incentive per patch as for a large change.

\noindent\textbf{The direction of change is inferred from masked context.}
Let $\mathcal{V}_{\ell}$ be the visible patches from which both predictions at $\ell$ are made. Under the shared partition $\mathbf{m}$, these are the positions on the other side of the partition from $\ell$, marked by dots in Fig.~\ref{fig:edds_concept}(a): $\mathbf{x}^{a}$ shows them in the mixed grid drawn there, and $\mathbf{x}^{b}$ shows them in the reciprocal grid, where its own patch $\ell$ is masked. Given $\mathcal{V}_{\ell}$, the direction $\boldsymbol{\delta}_{\ell}/\|\boldsymbol{\delta}_{\ell}\|_F$ remains uncertain and follows a distribution on the unit sphere. For $\varepsilon_{\mathrm{e}}=0$, the expected EDDS term at $\ell$ is minimized by any prediction $\hat{\boldsymbol{\delta}}_{\ell}$ pointing along
\begin{equation}
\mathbf{u}_{\ell}:=\mathbb{E}\!\left[\frac{\boldsymbol{\delta}_{\ell}}{\|\boldsymbol{\delta}_{\ell}\|_F}\,\middle|\,\mathcal{V}_{\ell},\ \ell\in\Gamma\right],
\label{eq:dime_edds_optimum}
\end{equation}
and its minimum is $1-\|\mathbf{u}_{\ell}\|_F$. Fig.~\ref{fig:edds_concept}(a) shows this geometry: EDDS charges $1-\cos\theta_{\ell}$ for the angle $\theta_{\ell}$ between the prediction and the true direction, and the optimal prediction points along the mean $\mathbf{u}_{\ell}$ of the distribution. The gain over predicting no change, $\|\mathbf{u}_{\ell}\|_F\in[0,1]$, is the length of this mean. It grows as the visible context determines the direction more consistently, and it does not depend on the magnitude of the change. Under $\mathcal{L}_{\mathrm{rgb}}$, the corresponding gain is the predictable variance of the residual, which shrinks with the size of the change. Because $\mathbf{u}_{\ell}$ is defined at every supported patch and must be inferred from the visible patches around it, EDDS supervises local, spatially resolved change, which favors dense features. This also suggests why DIME keeps benefiting from more data: once the shared appearance is well predicted, reconstruction offers little further gain, whereas each new same-identity pair still contributes changes whose directions remain to be predicted. Source isolation (Eq.~\eqref{eq:dime_source_attention_bias}) lets each prediction use only its own image, so the encoder must infer these directions from a single image, which is the representation transferred to downstream tasks.

\begin{table}[t]
\centering
\caption{\textbf{Architectures of DIME.} The decoder column reports its embedding dimension and block depth. The Params column counts the encoder only; the decoder and prediction heads, used only in pretraining, are excluded.}
\label{tab:dime_architecture}
\setlength{\tabcolsep}{4.0pt}
\vspace{-2pt}
\resizebox{\linewidth}{!}{%
\begin{tabular}{lccccc}
\toprule
Model &
Backbone &
Embed. dim &
Depth &
Decoder dim/depth &
Params \\
\midrule
DIME-S &
ViT-S/16 &
384 &
12 &
$384/6$ &
$23$M \\

DIME-B &
ViT-B/16 &
768 &
12 &
$512/8$ &
$86$M \\

DIME-L &
ViT-L/16 &
1024 &
24 &
$512/8$ &
$307$M \\
\bottomrule
\end{tabular}%
}
\vspace{-1.0em}
\end{table}

\noindent\textbf{EDDS rewards only predictable change.}
Same-identity pairs also differ in pose, lighting, and noise. EDDS ignores per-channel offsets and patch contrast changes (Sec.~\ref{sec:dime_losses}). If $\boldsymbol{\delta}_{\ell}$ is as likely as $-\boldsymbol{\delta}_{\ell}$ given $\mathcal{V}_{\ell}$, as for pure sensor noise, then $\mathbf{u}_{\ell}=\mathbf{0}$ and the expected gradient vanishes. Independent isotropic noise added to a change keeps the change's direction but discounts changes near the noise level. Predictable change, such as head rotation, is supervised like AU change within $\Gamma$, so EDDS rewards both without further separation.

\section{Experiments}

Sections~\ref{sec:exp_au}--\ref{sec:exp_generation} evaluate DIME on seven tasks: AU detection and intensity estimation, cross-domain face anti-spoofing, landmark localization, head pose estimation, face parsing, and face generation. Sec.~\ref{sec:exp_ablation} presents the ablation studies.

\subsection{Implementation Details and Experimental Settings}
\label{sec:implementation_settings}

\noindent\textbf{Architecture.}
We instantiate DIME with ViT encoders~\cite{dosovitskiy2020image,touvron2021training} using $16\times16$ patches and 2D rotary position embeddings (RoPE)~\cite{heo2024rotary}. Tokens of the two observations form the reciprocal grids of Eq.~\eqref{eq:dime_reciprocal_realization}, and every encoder block applies source isolation (Eq.~\eqref{eq:dime_source_attention_bias}). A lightweight MAE-style decoder~\cite{he2022masked} feeds two linear heads: one predicts normalized RGB patches for $\mathcal{L}_{\mathrm{rgb}}$, which also supply the reconstructions compared by $\mathcal{L}_{\mathrm{edds}}$, and the other predicts per-channel image gradients for $\mathcal{L}_{\mathrm{ori}}$. Table~\ref{tab:dime_architecture} lists detailed configurations.

\begin{table*}[t]
\centering
  \caption{Per-AU F1 scores (\%) on BP4D \cite{zhang2014bp4d}. Best and second-best results are in \BestScore{vermilion} and \SecondScore{deep blue}, respectively.}
  \label{tab:au_bp4d}
  \vspace{-1.0mm}

  \setlength{\tabcolsep}{2.5pt}
  \renewcommand{\arraystretch}{0.95}

  \resizebox{\textwidth}{!}{%
\begin{tabular}{@{}l*{14}{>{\centering\arraybackslash}m{2.9em}}@{}}
    \toprule
    Method
    & AU1 & AU2 & AU4 & AU6 & AU7 & AU10 & AU12
    & AU14 & AU15 & AU17 & AU23 & AU24
    & \textbf{Avg.}$\uparrow$ & \textbf{SD}$\downarrow$ \\
    \midrule

    CISNet \cite{chen2022causal}
    & 54.8 & 48.3 & 57.2 & 76.2 & 76.5
    & \SecondScore{85.2} & 87.2 & 66.2 & 50.9 & 65.0 & 47.7 & 56.5
    & 64.3 & 13.4 \\

    D-PAttNet \cite{onal2019d}
    & 50.7 & 42.5 & 59.0 & 79.4 & 79.0
    & 85.0 & 89.3 & 67.6 & 51.6 & 65.3 & 49.6 & 54.5
    & 64.5 & 14.9 \\

    CaFNet \cite{chen2021cafgraph}
    & 55.1 & 49.3 & 57.7 & 78.3 & 78.6
    & 85.1 & 86.2 & 67.4 & 52.0 & 64.4 & 48.3 & 56.2
    & 64.9 & 13.4 \\

    ME-GraphAU \cite{luo2022learning}
    & 52.7 & 44.3 & 60.9 & 79.9 & 80.1
    & \BestScore{85.3} & 89.2 & 69.4 & 55.4 & 64.4 & 49.8 & 55.1
    & 65.5 & 14.4 \\

    MAE-Face \cite{ma2024facial}
    & 65.2 & 55.7 & 66.9 & 79.5 & 79.5
    & \SecondScore{85.2} & \BestScore{90.0} & 67.5 & 59.7 & 67.5
    & 54.8 & 60.9
    & 69.4 & 11.1 \\

    FMAE \cite{ning2024representation}
    & 59.2 & 50.0 & 62.7 & 80.0 & 79.2
    & 84.7 & 89.8 & 63.5 & 52.8 & 65.1 & 55.3 & 56.9
    & 66.6 & 12.8 \\

    FMAE-IAT \cite{ning2024representation}
    & 62.7 & 51.9 & 62.7 & 79.8 & 80.1
    & 84.8 & \SecondScore{89.9} & 64.6 & 54.9 & 65.4 & 53.1 & 54.7
    & 67.1 & 12.7 \\
    \midrule

    MAE-L \cite{he2022masked}
    & 53.0 & 46.5 & 55.4 & 74.4 & 74.7
    & 83.4 & 85.4 & 64.4 & 49.1 & 63.2 & 45.9 & 54.7
    & 62.5 & 13.4 \\

    DINOv3-L \cite{simeoni2025dinov3}
    & 54.5 & 46.7 & 56.4 & 74.9 & 76.0
    & 83.2 & 85.5 & 66.0 & 49.0 & 64.0 & 47.7 & 54.6
    & 63.2 & 13.3 \\

    MixMAE-L \cite{liu2023mixmae}
    & 52.0 & 44.3 & 54.7 & 72.6 & 73.8
    & 83.6 & 85.5 & 62.9 & 46.7 & 62.1 & 43.9 & 53.6
    & 61.3 & 14.0 \\

    ConvNeXtV2-L \cite{woo2023convnext}
    & 52.8 & 45.6 & 55.7 & 73.9 & 74.6
    & 82.8 & 85.6 & 63.6 & 48.4 & 63.3 & 45.5 & 53.5
    & 62.1 & 13.6 \\
    \midrule

    DIME-S
    & 66.4 & 57.0 & 67.8 & 79.5 & 78.2
    & 84.2 & 88.2 & 67.7 & 61.6 & 68.2 & 56.4 & 62.4
    & 69.8 & 10.0 \\

    DIME-B
    & \SecondScore{68.3}
    & \SecondScore{61.6}
    & \SecondScore{69.4}
    & \SecondScore{80.1}
    & \BestScore{80.8}
    & 84.9
    & 89.2
    & \SecondScore{71.1}
    & \SecondScore{63.3}
    & \SecondScore{70.6}
    & \SecondScore{59.9}
    & \SecondScore{64.8}
    & \SecondScore{72.0}
    & \SecondScore{9.2} \\

    DIME-L
    & \BestScore{70.9}
    & \BestScore{62.3}
    & \BestScore{71.5}
    & \BestScore{81.3}
    & \SecondScore{80.4}
    & 84.9
    & 88.6
    & \BestScore{71.2}
    & \BestScore{66.4}
    & \BestScore{71.5}
    & \BestScore{63.0}
    & \BestScore{66.4}
    & \BestScore{\textbf{73.2}}
    & \BestScore{\textbf{8.3}} \\

    \bottomrule
  \end{tabular}%
  }
\vspace{-1.5em}
\end{table*}

\noindent\textbf{Pretraining and data.}
We pretrain DIME on Face-20M, comprising 20M facial images aggregated from WebFace42M~\cite{zhu2023webface260m}, Glint360K~\cite{an2022killing}, and MS1MV2~\cite{deng2022arcface}. Identity labels are inherited from these sources and are used only to sample pairs (Sec.~\ref{sec:dime_formulation}); pretraining uses no other annotations. The original images are restored to $448\times448$ and $512\times512$ using GFPGAN~\cite{wang2021towards} and CodeFormer~\cite{zhou2022towards}.

Following the recipe of Hiera~\cite{ryali2023hiera}, we use AdamW with $(\beta_1,\beta_2)=(0.9,0.95)$ and weight decay $0.05$. The pretraining stage uses a 40-epoch warmup followed by cosine decay. Training runs on the EU LUMI supercomputer with AMD MI250X GPUs. We first train all DIME variants at $224\times224$ for 400 epochs, then continue at $512\times512$ for another 100 epochs. With the patch size fixed, this transition increases the spatial grid from $14\times14$ to $32\times32$ tokens, allowing continued pretraining to resolve finer details. The complete training configurations are provided with the released code.

\noindent\textbf{Pretraining hyperparameters.}
Each minibatch pairs every image with another image of the same identity. We use $r=0.5$, $\lambda_{\mathrm{edds}}=1.0$, and $\lambda_{\mathrm{ori}}=0.5$. EDDS and $\mathcal{L}_{\mathrm{ori}}$ share $\rho=0.5$, so each supervises the upper half of the target residuals or gradients by magnitude, and the pair gate uses $\kappa=0.2$. EDDS is enabled after a 20-epoch warm-up, so that the differences it compares come from already formed reconstructions; we use $\varepsilon_{\mathrm{n}}=10^{-6}$. The base learning rate is $8\times10^{-5}$ per 256 images and scales linearly with the batch size. We clip gradients at norm 5.0, use bfloat16 mixed precision, and apply no dropout or stochastic depth.

\noindent\textbf{Transfer.}
After pretraining, we retain only the ViT encoder for downstream tasks. Token mixing and the source-isolation mask are removed, so the encoder processes single images exactly as a standard ViT. The encoder is fine-tuned end to end with each task head, except for the evaluated face generation task and the frozen probes of Table~\ref{tab:dime_ablation}(e).

\noindent\textbf{Pretraining cost.}
One DIME epoch iterates over disjoint same-identity pairs that cover every image once, as in MAE. In contrast, MAE removes a different random subset of tokens from every image, which requires per-sample gather and scatter operations and leaves the encoder with short sequences of $25\%$ of the tokens that use the accelerator inefficiently. DIME encodes fixed-length sequences in which each position's source is selected by the mask, and source isolation is computed as attention within each of the two groups, which halves the attention cost of a full sequence. With the same batch size, precision, and hardware (AMD MI250X GPUs), DIME-B processes 493 images per second per GPU, compared with 422 for MAE-B. DIME is also trained for fewer epochs than common MAE recipes: 400 epochs at $224\times224$ and 100 at $512\times512$, compared with 1600 epochs for MAE~\cite{he2022masked} and 800 for MAE-Face~\cite{ma2024facial}. In total, pretraining DIME-B took $\sim$4500 GPU hours, compared with $\sim$5300 for MAE-B under the schedule of Fig.~\ref{fig:1}(d).

\noindent\textbf{Comparison protocols.} For the generic visual models MAE-L~\cite{he2022masked}, DINOv2-L~\cite{oquab2023dinov2}, DINOv3-L~\cite{simeoni2025dinov3}, MixMAE-L~\cite{liu2023mixmae}, and ConvNeXtV2-L~\cite{woo2023convnext}, we start from the official pretrained weights and apply the same downstream protocol as DIME in Tables~\ref{tab:au_bp4d}--\ref{tab:cross_domain_fas} and~\ref{tab:face_generation}. These checkpoints differ from DIME in pretraining data, resolution, and compute; the matched comparisons with MAE on Face-20M (Fig.~\ref{fig:1}(d) and Table~\ref{tab:dime_ablation}) isolate the effect of the objective. For task-specific methods, we report the results published in the original papers.

\begin{table*}[!t]
\centering
  \caption{Per-AU F1 scores (\%) on BP4D+ \cite{zhang2016multimodal}. Best and second-best results are in \BestScore{vermilion} and \SecondScore{deep blue}, respectively.}
  \label{tab:au_bp4dplus}
  \vspace{-1.0mm}

  \setlength{\tabcolsep}{2.5pt}
  \renewcommand{\arraystretch}{0.95}

  \resizebox{\textwidth}{!}{%
\begin{tabular}{@{}l*{14}{>{\centering\arraybackslash}m{2.9em}}@{}}
    \toprule
    Method
    & AU1 & AU2 & AU4 & AU6 & AU7 & AU10 & AU12
    & AU14 & AU15 & AU17 & AU23 & AU24
    & \textbf{Avg.}$\uparrow$ & \textbf{SD}$\downarrow$ \\
    \midrule

    FACS3D-Net \cite{yang2019facs3d}
    & 43.0 & 38.1 & 49.9 & 82.3 & 85.1 & 87.2 & 87.5
    & 66.0 & 48.4 & 47.4 & 50.0 & 31.9
    & 59.7 & 19.8 \\

    ML-GCN \cite{chen2019multi}
    & 40.2 & 36.9 & 32.5 & 84.8 & \SecondScore{88.9}
    & 89.6 & 89.3 & 81.2 & 53.3 & 43.1 & 55.9 & 28.3
    & 60.3 & 23.6 \\

    MS-CAM \cite{you2020cross}
    & 38.3 & 37.6 & 25.0 & 85.0 & \BestScore{90.9}
    & 90.9 & 89.0 & 81.5 & \BestScore{60.9} & 40.6 & 58.2 & 28.0
    & 60.5 & 24.9 \\

    SEV-Net \cite{yang2021exploiting}
    & 47.9 & 40.8 & 31.2 & 86.9 & 87.5
    & 89.7 & 88.9 & 82.6 & 39.9 & 55.6 & 59.4 & 27.1
    & 61.5 & 23.3 \\

    MAE-Face \cite{ma2024facial}
    & 53.0 & 44.5 & 49.4 & \BestScore{87.7} & 88.8
    & \BestScore{91.4} & \BestScore{90.8} & 83.2 & 45.2 & 50.0
    & 61.2 & 45.3
    & 65.9 & 19.6 \\

    FMAE \cite{ning2024representation}
    & 53.9 & 45.5 & 45.9 & 86.2 & 88.3
    & \SecondScore{91.2} & 89.9 & 82.3 & 51.3 & 56.3 & 60.7 & 42.7
    & 66.2 & 18.8 \\

    FMAE-IAT \cite{ning2024representation}
    & 54.2 & 47.0 & 53.9 & 85.7 & 88.4
    & \SecondScore{91.2} & 89.7 & 82.4 & 50.3 & 54.4 & 61.0 & 43.4
    & 66.8 & 18.1 \\
    \midrule

    MAE-L \cite{he2022masked}
    & 46.2 & 39.0 & 29.4 & 85.1 & 85.7
    & 88.2 & 87.1 & 80.8 & 38.1 & 53.5 & 57.6 & 25.3
    & 59.7 & 23.4 \\

    DINOv3-L \cite{simeoni2025dinov3}
    & 47.6 & 38.8 & 31.2 & 85.4 & 86.7
    & 87.9 & 87.3 & 82.3 & 38.0 & 54.3 & 59.2 & 25.6
    & 60.4 & 23.3 \\

    MixMAE-L \cite{liu2023mixmae}
    & 44.5 & 38.2 & 27.0 & 84.0 & 85.0
    & 88.3 & 87.3 & 79.2 & 35.8 & 52.6 & 55.8 & 23.8
    & 58.5 & 23.9 \\

    ConvNeXtV2-L \cite{woo2023convnext}
    & 45.8 & 39.5 & 28.4 & 84.9 & 86.0
    & 87.3 & 87.7 & 80.1 & 36.9 & 53.9 & 56.8 & 23.8
    & 59.3 & 23.7 \\
    \midrule

    DIME-S
    & 55.9 & 47.0 & 52.2 & 86.8 & 86.9
    & 90.0 & 89.0 & 82.1 & 48.8 & 52.1 & 62.5 & 48.3
    & 66.8 & 17.5 \\

    DIME-B
    & \SecondScore{58.4}
    & \SecondScore{52.9}
    & \SecondScore{55.3}
    & \SecondScore{87.5}
    & 88.7
    & 90.1
    & \SecondScore{90.1}
    & \BestScore{84.3}
    & 51.6
    & \SecondScore{57.0}
    & \SecondScore{65.6}
    & \SecondScore{52.5}
    & \SecondScore{69.5}
    & \SecondScore{16.2} \\

    DIME-L
    & \BestScore{62.1}
    & \BestScore{54.2}
    & \BestScore{59.0}
    & 87.1
    & 88.1
    & 90.3
    & 89.5
    & \SecondScore{83.5}
    & \SecondScore{56.2}
    & \BestScore{58.5}
    & \BestScore{68.0}
    & \BestScore{55.5}
    & \BestScore{\textbf{71.0}}
    & \BestScore{\textbf{14.6}} \\

    \bottomrule
  \end{tabular}%
  }
\vspace{-0.5em}
\end{table*}

\begin{table*}[!t]
\centering
\begin{minipage}[t]{\dimexpr(\textwidth-\columnsep)/2\relax}
\centering
  \caption{AU detection on DISFA~\cite{mavadati2013disfa}. Best and second-best results are in \BestScore{vermilion} and \SecondScore{deep blue}, respectively.}
  \label{tab:au_disfa}
  \vspace{-1.0mm}
  \setlength{\tabcolsep}{2.0pt}
  \renewcommand{\arraystretch}{1.02}
  \resizebox{\linewidth}{!}{%
  \begin{tabular}{@{}l*{10}{c}@{}}
    \toprule
    Method
    & AU1 & AU2 & AU4 & AU6 & AU9 & AU12 & AU25 & AU26
    & \textbf{Avg.}$\uparrow$ & \textbf{SD}$\downarrow$ \\
    \midrule

    CISNet \cite{chen2022causal}
    & 48.8 & 50.4 & \SecondScore{78.9} & 51.9 & 47.1
    & \SecondScore{80.1} & 95.4 & 65.0
    & 64.7 & 17.0 \\

    CaFNet \cite{chen2021cafgraph}
    & 45.6 & 55.7 & \BestScore{80.2} & 51.0 & 54.7
    & 79.0 & 95.2 & 65.3
    & 65.8 & 16.2 \\

    MGRR-Net \cite{ge2024mgrr}
    & 61.3 & 62.9 & 75.8 & 48.7 & 53.8
    & 75.5 & 94.3 & 73.1
    & 68.2 & 13.6 \\

    MAE-Face \cite{ma2024facial}
    & 72.0 & 68.7 & 75.2 & 59.8 & 58.5
    & 79.3 & 96.4 & 73.6
    & 72.9 & 11.2 \\

    FMAE-IAT \cite{ning2024representation}
    & 64.7 & 61.3 & 70.8 & 58.1 & 59.4
    & 79.9 & 95.2 & 71.3
    & 70.1 & 11.7 \\
    \midrule

    MAE-L \cite{he2022masked}
    & 50.7 & 50.5 & 77.6 & 52.0 & 47.2
    & 79.8 & 95.5 & 65.1
    & 64.8 & 16.6 \\

    DINOv3-L \cite{simeoni2025dinov3}
    & 47.7 & 48.1 & 78.0 & 49.2 & 46.2
    & 79.7 & 93.0 & 62.1
    & 63.0 & 17.1 \\

    MixMAE-L \cite{liu2023mixmae}
    & 46.6 & 48.9 & 77.4 & 48.7 & 44.4
    & 79.8 & 92.4 & 61.8
    & 62.5 & 17.2 \\

    ConvNeXtV2-L \cite{woo2023convnext}
    & 48.1 & 49.4 & 78.2 & 50.3 & 46.8
    & 79.9 & 93.7 & 63.2
    & 63.7 & 16.9 \\
    \midrule

    DIME-S
    & 72.2 & 69.6 & 75.4 & 60.8 & 59.8
    & 79.1 & 96.3 & \SecondScore{74.0}
    & 73.4 & 10.7 \\

    DIME-B
    & \SecondScore{72.6}
    & \SecondScore{70.5}
    & 75.9
    & \SecondScore{61.6}
    & \SecondScore{60.5}
    & 79.2
    & \SecondScore{96.5}
    & \BestScore{74.4}
    & \SecondScore{73.9}
    & \SecondScore{10.5} \\

    DIME-L
    & \BestScore{73.2}
    & \BestScore{71.7}
    & 76.6
    & \BestScore{62.8}
    & \BestScore{61.8}
    & \BestScore{80.2}
    & \BestScore{97.4}
    & 73.9
    & \BestScore{\textbf{74.7}}
    & \BestScore{\textbf{10.4}} \\

    \bottomrule
  \end{tabular}%
  }
\end{minipage}\hfill%
\begin{minipage}[t]{\dimexpr(\textwidth-\columnsep)/2\relax}
\vspace{0pt}
\centering
  \caption{AU intensity estimation. Best and second-best results are in \BestScore{vermilion} and \SecondScore{deep blue}, respectively.}
  \label{tab:au_intensity}
  \vspace{-1.0mm}
  \setlength{\tabcolsep}{16pt}
  \renewcommand{\arraystretch}{0.92}
  \resizebox{\linewidth}{!}{%
  \begin{tabular}{@{}lcc cc@{}}
    \toprule
    & \multicolumn{2}{c}{BP4D \cite{zhang2014bp4d}} & \multicolumn{2}{c}{DISFA \cite{mavadati2013disfa}} \\
    \cmidrule(lr){2-3}\cmidrule(l){4-5}
    Method
    & ICC$\uparrow$ & MSE$\downarrow$
    & ICC$\uparrow$ & MSE$\downarrow$ \\
    \midrule
    ISIR \cite{nicolle2015facial}           & .720 & .850 & --    & --    \\
    HR \cite{ntinou2021transfer}            & .730 & .780 & .570  & .320  \\
    APs \cite{sanchez2021affective}           & .740 & .770 & .580  & .300  \\
    MAE-Face \cite{ma2024facial}       & .742 & .794 & .674  & .213  \\
    \midrule
    MAE-L \cite{he2022masked}          & .700 & .860 & .640  & .270  \\
    DINOv3-L \cite{simeoni2025dinov3}       & .710 & .840 & .610  & .280  \\
    MixMAE-L \cite{liu2023mixmae}       & .690 & .880 & .610  & .290  \\
    ConvNeXtV2-L \cite{woo2023convnext}  & .700 & .850 & .620  & .280  \\
    \midrule
    DIME-S
    & .761 & .752 & .693 & .208 \\
    DIME-B
    & \SecondScore{.768} & \SecondScore{.742}
    & \SecondScore{.698} & \SecondScore{.196} \\
    DIME-L
    & \BestScore{\textbf{.774}} & \BestScore{\textbf{.735}}
    & \BestScore{\textbf{.702}} & \BestScore{\textbf{.194}} \\
    \bottomrule
  \end{tabular}%
  }
\end{minipage}
\vspace{-1.0em}
\end{table*}

\begin{figure*}[!t]
\centering
    \setlength{\abovecaptionskip}{2pt}
    \includegraphics[width=\linewidth]{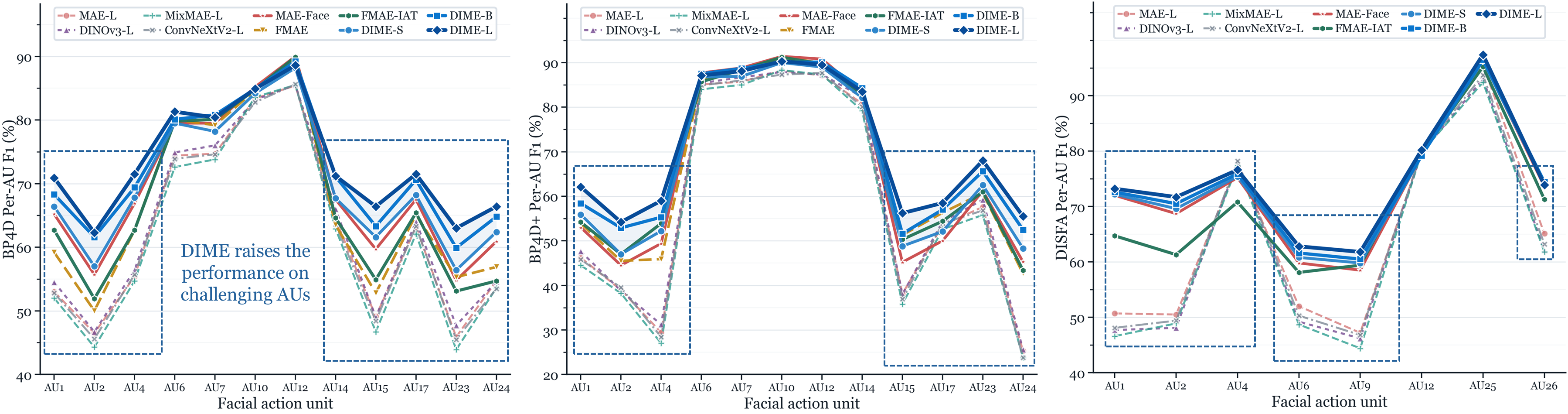}
    \caption{Per-AU F1 scores on BP4D (left), BP4D+ (middle), and DISFA (right). Dashed boxes highlight groups containing challenging AUs, where DIME improves detection beyond generic visual encoders and prior facial self-supervised models.}
    \label{fig:au_detection_analysis}
    \vspace{-1.5em}
\end{figure*}

\subsection{Experiments on Facial Action Unit Understanding}
\label{sec:exp_au}

Facial action units (AUs) provide an objective, anatomically grounded description of facial muscle activity. AU detection and intensity estimation are challenging: each AU induces only subtle changes that are easily overwhelmed by identity, pose, and illumination, and many AUs co-occur with imbalanced frequencies. They therefore demand representations that are sensitive to fine-grained facial variation.

\noindent\textbf{Evaluation protocol.} Following MAE-Face~\cite{ma2024facial}, we evaluate AU detection on BP4D~\cite{zhang2014bp4d}, BP4D+~\cite{zhang2016multimodal}, and DISFA~\cite{mavadati2013disfa}, containing 12, 12, and 8 AUs from 41, 140, and 27 subjects, respectively. Newly evaluated encoders use the subject-exclusive three-fold splits of~\cite{ma2024facial}, with patch tokens processed by global average pooling, layer normalization, and a linear head trained using inverse-frequency-weighted binary cross-entropy. We report per-AU F1, its unweighted mean, and the population SD across AUs; SD measures performance balance across AUs and is interpreted jointly with mean F1. For intensity estimation, we use the FERA2015 train/dev partition \cite{valstar2015fera} of BP4D with five AUs and subject-exclusive three-fold evaluation on DISFA with twelve AUs, following the regression objective of MAE-Face and reporting mean ICC(3,1) and MSE. Data processing and optimization settings are held fixed across newly evaluated encoders.

\noindent\textbf{AU detection on BP4D.} Table~\ref{tab:au_bp4d} reports per-AU F1, mean F1, and cross-AU SD. Among established methods, MAE-Face achieves 69.4\% mean F1, compared with 66.6\% for FMAE and 67.1\% for FMAE-IAT. DIME-S reaches 69.8\%, while DIME-B and DIME-L further improve to 72.0\% and 73.2\%. The best generic encoder, DINOv3-L, trails DIME-S by 6.6 points (63.2\%). SD decreases from 11.1 for MAE-Face to 10.0, 9.2, and 8.3 for the three DIME variants, indicating increasingly balanced performance across AUs. AU12 remains among the strongest categories, whereas AU2 and AU23 remain challenging. Overall, the joint improvement in mean F1 and SD suggests that DIME benefits a broader range of AUs rather than only strong categories.

\noindent\textbf{AU detection on BP4D+.} On BP4D+ (Table~\ref{tab:au_bp4dplus}), DIME-S, DIME-B, and DIME-L reach 66.8\%, 69.5\%, and 71.0\%, compared with 65.9\% for MAE-Face, and lower the cross-AU SD from 19.6 to 17.5, 16.2, and 14.6. From DIME-B to DIME-L, mean F1 rises by 1.5 points; the largest gains are on AU15, AU1, AU4, AU24, and AU23, while DIME-B is higher on AU6, AU7, AU12, and AU14. On the seven AUs below 66\% for DIME-B, DIME-L gains 2.9 points on average, whereas the five above 84\% each change by under one point.

\noindent\textbf{AU detection on DISFA.} DISFA (Table~\ref{tab:au_disfa}) has fewer subjects than BP4D and BP4D+ and a strongly imbalanced AU distribution. DIME-S, DIME-B, and DIME-L reach 73.4\%, 73.9\%, and 74.7\% mean F1, compared with 72.9\% for MAE-Face, and lower the cross-AU SD from 11.2 to 10.7, 10.5, and 10.4. These gains are smaller than on BP4D+. AU25 remains the easiest AU, and AU6 and AU9 remain the hardest.

\begin{table*}[!t]
  \centering
  \caption{Cross-domain face anti-spoofing results. The generic visual encoders and DIME variants are evaluated using a unified downstream protocol \cite{yu2025learning}. Best and second-best results are highlighted in \BestScore{vermilion} and \SecondScore{deep blue}, respectively.}
  \label{tab:cross_domain_fas}
  \vspace{-1.0mm}
  \setlength{\tabcolsep}{10pt}
  \renewcommand{\arraystretch}{0.94}
  \resizebox{\textwidth}{!}{%
  \begin{tabular}{@{}l*{5}{cc}@{}}
    \toprule
    & \multicolumn{2}{c}{$ICM\!\rightarrow\!O$}
    & \multicolumn{2}{c}{$OCM\!\rightarrow\!I$}
    & \multicolumn{2}{c}{$OCI\!\rightarrow\!M$}
    & \multicolumn{2}{c}{$OMI\!\rightarrow\!C$}
    & \multicolumn{2}{c}{Average} \\
    \cmidrule(lr){2-3}
    \cmidrule(lr){4-5}
    \cmidrule(lr){6-7}
    \cmidrule(lr){8-9}
    \cmidrule(l){10-11}
    Method
    & HTER$\downarrow$ & AUC$\uparrow$
    & HTER$\downarrow$ & AUC$\uparrow$
    & HTER$\downarrow$ & AUC$\uparrow$
    & HTER$\downarrow$ & AUC$\uparrow$
    & HTER$\downarrow$ & AUC$\uparrow$ \\
    \midrule

    LBP-TOP \cite{zhao2007dynamic}
    & 30.17 & 77.61
    & 29.14 & 71.69
    & 36.90 & 70.80
    & 33.52 & 73.15
    & 32.43 & 73.31 \\

    SA-FAS \cite{sun2023rethinking}
    & 10.00 & 96.23
    & 6.58 & 97.54
    & 5.95 & 96.55
    & 8.78 & 95.37
    & 7.83 & 96.42 \\

    PDLE \cite{kwak2023liveness}
    & 11.42 & 95.52
    & 8.62 & 97.60
    & 5.41 & 98.85
    & 10.05 & 94.27
    & 8.88 & 96.56 \\

    UDG-FAS \cite{liu2023towards}
    & 10.97 & 95.36
    & 5.86 & 98.62
    & 5.95 & 98.47
    & 9.82 & \SecondScore{96.76}
    & 8.15 & 97.30 \\

    GAC-FAS \cite{le2024gradient}
    & 8.60 & 97.16
    & 4.29 & 98.87
    & 5.00 & 97.56
    & 8.20 & 95.16
    & 6.52 & 97.19 \\

    BIB-FAS \cite{yu2025learning}
    & 7.47 & \SecondScore{97.77}
    & 3.95 & 99.11
    & 4.05 & 98.27
    & 7.42 & 96.60
    & 5.72 & 97.94 \\
    \midrule

    MAE-L \cite{he2022masked}
    & 18.39 & 90.52
    & 14.17 & 92.15
    & 12.45 & 91.68
    & 15.58 & 90.21
    & 15.15 & 91.14 \\

    MixMAE-L \cite{liu2023mixmae}
    & 17.61 & 91.11
    & 15.39 & 92.90
    & 15.17 & 89.82
    & 19.70 & 87.73
    & 16.97 & 90.39 \\

    DINOv3-L \cite{simeoni2025dinov3}
    & 15.30 & 92.23
    & 13.37 & 94.26
    & 14.06 & 90.79
    & 16.44 & 89.96
    & 14.79 & 91.81 \\

    ConvNeXtV2-L \cite{woo2023convnext}
    & 16.57 & 88.75
    & 14.33 & 93.35
    & 13.07 & 91.31
    & 17.40 & 88.35
    & 15.34 & 90.44 \\
    \midrule

    DIME-S
    & 7.41 & 97.68
    & \SecondScore{3.93} & \SecondScore{99.19}
    & \SecondScore{3.91} & 98.88
    & 7.07 & 96.53
    & 5.58 & 98.07 \\

    DIME-B
    & \SecondScore{7.38} & 97.73
    & \BestScore{3.88} & \BestScore{99.23}
    & \BestScore{3.82} & \BestScore{98.97}
    & \SecondScore{6.93} & 96.62
    & \SecondScore{5.50} & \SecondScore{98.14} \\

    DIME-L
    & \BestScore{7.13} & \BestScore{97.92}
    & 3.94 & 99.11
    & 3.94 & \SecondScore{98.92}
    & \BestScore{6.71} & \BestScore{97.43}
    & \BestScore{\textbf{5.43}} & \BestScore{\textbf{98.35}} \\

    \bottomrule
  \end{tabular}%
  }
  \vspace{-1.5em}
\end{table*}

\noindent\textbf{Performance on challenging AUs.}
Fig.~\ref{fig:au_detection_analysis} compares per-AU F1 on BP4D, BP4D+, and DISFA from left to right. DIME's gains concentrate on lower-performing AUs highlighted by the dashed boxes. Compared with MAE-Face, DIME-L improves AU2, AU15, and AU23 by 6.6, 6.7, and 8.2 points on BP4D, and AU2, AU4, and AU24 by 9.7, 9.6, and 10.2 points on BP4D+. High-performing categories such as AU10 and AU12 remain broadly comparable. These improvements support more effective modeling of discriminative facial changes across challenging categories.

\noindent\textbf{AU intensity estimation.} Table~\ref{tab:au_intensity} tests whether DIME transfers from binary AU occurrence to continuous intensity. DIME-S already outperforms MAE-Face on both datasets and metrics. On BP4D, DIME-L raises ICC from .742 to .774 and reduces MSE from .794 to .735; on DISFA, it raises ICC from .674 to .702 and reduces MSE from .213 to .194. The small gains from DIME-S to DIME-L suggest that the objective matters more than encoder size.

\noindent\textbf{Parameter efficiency.}
Fig.~\ref{fig:1}(c) compares mean detection F1 over BP4D and BP4D+ against model size. DIME-S achieves 68.3\% with 23M parameters, exceeding MAE-Face's mean F1 while using 73\% fewer parameters. DIME-B and DIME-L further improve accuracy with increased capacity.

\noindent\textbf{Scaling analysis.}
Fig.~\ref{fig:1}(d) compares MAE-B and DIME-B pretrained on Face-20M fractions from 10\% to 100\%, with the same encoder, data, training schedule, and downstream evaluation; DIME-B also uses about 15\% fewer GPU hours (Sec.~\ref{sec:implementation_settings}). From 6M to 20M images, MAE-B improves only from 64.5\% to about 65.0\% BP4D mean F1, whereas DIME-B rises from 66.1\% to 72.0\%. This matches Sec.~\ref{sec:dime_formal}: once the shared appearance is well predicted, reconstruction gains little from more images, whereas each new same-identity pair still adds changes whose directions EDDS must predict.

\subsection{Experiments on Cross-Domain Face Anti-Spoofing}
\label{sec:exp_cross_domain_fas}

Face anti-spoofing (FAS)~\cite{yu2025learning} protects face-based biometric systems from presentation attacks. The cues that separate live from spoof faces can be subtle and vary with sensor, illumination, identity, and attack medium, so models trained on fixed datasets may exploit domain-specific appearance or acquisition artifacts that are absent at deployment. Cross-domain FAS tests whether a representation captures evidence that transfers to unseen datasets.

\noindent\textbf{Protocol and metrics.} Following BIB-FAS~\cite{yu2025learning}, we adopt a leave-one-out protocol over four publicly available datasets: CASIA-FASD~\cite{zhang2012face} ($C$), MSU-MFSD~\cite{wen2015face} ($M$), OULU-NPU~\cite{boulkenafet2017oulu} ($O$), and Idiap Replay-Attack~\cite{chingovska2012effectiveness} ($I$). Each model is trained on three source datasets and evaluated on one unseen dataset, which produces four experimental settings: $ICM\!\rightarrow\!O$, $OCM\!\rightarrow\!I$, $OCI\!\rightarrow\!M$, and $OMI\!\rightarrow\!C$.

We report Half Total Error Rate (HTER), the average of false-acceptance and false-rejection rates, and area under the receiver operating characteristic curve (AUC); lower HTER and higher AUC are better. For specialized FAS methods, we report their official results. For newly evaluated models, checkpoints and HTER decision thresholds are selected exclusively using held-out source-domain validation data.

\noindent\textbf{Results.} Table~\ref{tab:cross_domain_fas} compares DIME with specialized FAS methods and generic encoders. Among the specialized methods, BIB-FAS achieves the strongest average performance, with 5.72\% HTER and 97.94\% AUC. Under the unified protocol, DIME-S already exceeds it with 5.58\% HTER and 98.07\% AUC. DIME-B reaches 5.50\% HTER and 98.14\% AUC, and DIME-L achieves the best averages of 5.43\% HTER and 98.35\% AUC. DIME-L also matches or exceeds every specialized method on both metrics in all four settings, most clearly on $OMI\!\rightarrow\!C$ (6.71\% vs.\ 7.42\% HTER).

\noindent\textbf{Comparison with generic pretraining.} DINOv3-L obtains 14.79\% HTER and 91.81\% AUC, whereas DIME-S improves these metrics by 9.21 and 6.26 points. The remaining generic encoders achieve 15.15--16.97\% HTER and 90.39--91.14\% AUC. In particular, MAE-L reaches 15.15\% HTER and 91.14\% AUC, compared with 5.43\% and 98.35\% for DIME-L. Part of this gap comes from face-domain data, yet the objective also matters: on the same images, the full DIME objective lowers HTER from 11.32\% for single-image MAE to 8.11\%, with EDDS contributing the largest share (Table~\ref{tab:dime_ablation}). This is consistent with single-image reconstruction emphasizing static appearance and acquisition statistics, while DIME prioritizes identity-controlled facial variation that transfers more reliably across domains.

\subsection{Experiments on Facial Geometry Estimation}
\label{sec:exp_geometry}

We evaluate DIME's transferability to geometry-sensitive tasks on three levels: sparse geometry through facial landmark detection \cite{zhou2023star}, global 3D orientation through head pose estimation \cite{hempel20226d}, and dense geometry through face parsing \cite{te2021agrnet}. Since these tasks primarily depend on facial configuration and organization rather than identity-level knowledge, they provide a direct test of whether DIME's pretraining captures facial representations beyond identity.

\begin{table*}[t]
  \centering
  \setlength{\abovecaptionskip}{2pt}

  \def\GeometryTableFormat{%
    \fontsize{7.6pt}{8.4pt}\selectfont
    \setlength{\tabcolsep}{0.25pt}%
    \renewcommand{\arraystretch}{1.1}%
  }

  \caption{Facial geometry estimation: sparse geometry (landmark detection), global orientation (head pose estimation), and dense semantic structure (face parsing). Best and second-best results are in \BestScore{vermilion} and \SecondScore{deep blue}.}
  \label{tab:facial_geometry}

  \begin{minipage}[t]{0.250\textwidth}
    \vspace{0pt}
    \centering
    \GeometryTableFormat

    \begin{tabular*}{\linewidth}{
      @{\extracolsep{\fill}}lccc@{}
    }
      \toprule
      & \multicolumn{3}{c}{WFLW \cite{wu2018look}} \\
      \cmidrule(l){2-4}
      Method
      & NME$\downarrow$
      & FR$\downarrow$
      & AUC$\uparrow$ \\
      \midrule
      HRNet-W18 \cite{wang2020deep}
      & 4.60 & -- & -- \\
      STAR \cite{zhou2023star}
      & 4.02 & 2.32 & 60.50 \\
      ADNet \cite{huang2021adnet}
      & 4.14 & 2.72 & 60.22 \\
      LDEQ \cite{micaelli2023recurrence}
      & 3.92 & 2.48 & 62.40 \\
      Faceptor-Full \cite{qin2024faceptor}
      & 4.03 & 1.92 & 60.24 \\
      FaRL-448-ft \cite{zheng2022general}
      & 3.96 & \SecondScore{1.76} & 61.16 \\
      ORFormer \cite{chiang2025orformer}
      & 3.86 & \SecondScore{1.76} & 62.20 \\
      \midrule
      DIME-S
      & 3.93 & \SecondScore{1.76} & 61.28 \\
      DIME-B
      & \SecondScore{3.82}
      & \BestScore{\textbf{1.72}}
      & \SecondScore{62.94} \\
      DIME-L
      & \BestScore{\textbf{3.74}}
      & \BestScore{\textbf{1.72}}
      & \BestScore{\textbf{63.74}} \\
      \bottomrule
    \end{tabular*}

    \vspace{0.7mm}
    \parbox[t][1.55cm][t]{\linewidth}{%
      \footnotesize
      \textbf{(a) Landmark detection.}
      Inter-ocular NME (\%), FR (\%), and AUC (\%) on the WFLW test set; FR and AUC use an NME threshold of $0.10$.
    }
  \end{minipage}
  \hfill
  \begin{minipage}[t]{0.425\textwidth}
    \vspace{0pt}
    \centering
    \GeometryTableFormat

    \begin{tabular*}{\linewidth}{
      @{\extracolsep{\fill}}lcccccccc@{}
    }
      \toprule
      & \multicolumn{4}{c}{AFLW2000-3D \cite{zhu2019face}}
      & \multicolumn{4}{c}{BIWI \cite{fanelli2013random}} \\
      \cmidrule(lr){2-5}\cmidrule(l){6-9}
      Method
      & Yaw & Pitch & Roll & Mean
      & Yaw & Pitch & Roll & Mean \\
      \midrule
      TriNet \cite{cao2021vector}
      & 4.36 & 5.81 & 4.51 & 4.89
      & \BestScore{\textbf{3.11}} & 5.09 & 5.20 & 4.47 \\
      MNN \cite{valle2020multi}
      & 3.34 & 4.69 & 3.48 & 3.83
      & 3.98 & 4.61 & \BestScore{\textbf{2.39}} & 3.66 \\
      RankPose \cite{dai2020rankpose}
      & 3.26 & 4.72 & 3.23 & 3.74
      & 4.54 & 5.61 & 3.05 & 4.40 \\
      img2pose \cite{albiero2021img2pose}
      & 3.42 & 5.03 & 3.28 & 3.91
      & 4.56 & \BestScore{\textbf{3.54}} & 3.25 & 3.78 \\
      6DRepNet \cite{hempel20226d}
      & 3.27 & 4.58 & 2.98 & 3.61
      & 3.23 & 5.32 & 2.78 & 3.78 \\
      SRHP-Euler \cite{cobo2024representation}
      & 2.76 & 4.25 & 2.76 & 3.26
      & 4.54 & 5.05 & 2.80 & 4.13 \\
      SRHP-6D \cite{cobo2024representation}
      & 2.85 & 4.59 & 3.04 & 3.49
      & 4.58 & 4.65 & 2.71 & 3.98 \\
      \midrule
      DIME-S
      & 2.82 & \SecondScore{3.86} & 2.71 & 3.13
      & 3.47 & 4.26 & 2.62 & 3.45 \\
      DIME-B
      & \BestScore{\textbf{2.56}}
      & \BestScore{\textbf{3.82}}
      & \BestScore{\textbf{2.65}}
      & \BestScore{\textbf{3.01}}
      & 3.22 & 4.17 & 2.54
      & \SecondScore{3.31} \\
      DIME-L
      & \SecondScore{2.70}
      & \SecondScore{3.86}
      & \SecondScore{2.68}
      & \SecondScore{3.08}
      & \SecondScore{3.21}
      & \SecondScore{4.07}
      & \SecondScore{2.53}
      & \BestScore{\textbf{3.27}} \\
      \bottomrule
    \end{tabular*}

    \vspace{0.7mm}
    \parbox[t][1.55cm][t]{\linewidth}{%
      \footnotesize
      \textbf{(b) Head pose estimation.}
      Models are trained on 300W-LP and evaluated on AFLW2000-3D and BIWI.
      Results are mean absolute angular errors in degrees (lower is better); Mean averages yaw, pitch, and roll over all test images.
    }
  \end{minipage}
  \hfill
  \begin{minipage}[t]{0.290\textwidth}
    \vspace{0pt}
    \centering
    \GeometryTableFormat

    \begin{tabular*}{\linewidth}{
      @{\extracolsep{\fill}}lcccc@{}
    }
      \toprule
      & \multicolumn{2}{c}{LaPa \cite{liu2020new}}
      & \multicolumn{2}{c}{CelebM \cite{lee2020maskgan}} \\
      \cmidrule(lr){2-3}\cmidrule(l){4-5}
      Method
      & F1$\uparrow$ & mIoU$\uparrow$
      & F1$\uparrow$ & mIoU$\uparrow$ \\
      \midrule
      AGRNet \cite{te2021agrnet}
      & 92.30 & 87.00 & 85.50 & -- \\
      FP-LIIF \cite{sarkar2023parameter}
      & 92.41 & -- & 86.14 & -- \\
      DML-CSR \cite{zheng2022decoupled}
      & 92.38 & 87.13 & 86.07 & 77.81 \\
      FaRL-448-ft \cite{zheng2022general}
      & 93.88 & -- & 89.56 & -- \\
      SegFace-512 \cite{narayan2025segface}
      & 93.03 & 88.14 & 88.96 & 81.55 \\
      PaCo-FR \cite{xie2026paco}
      & 93.91 & -- & 89.63 & -- \\
      Faceptor-Full \cite{qin2024faceptor}
      & 91.50 & -- & 88.20 & -- \\
      \midrule
      DIME-S
      & 93.86 & 88.96 & 89.61 & 81.96 \\
      DIME-B
      & \SecondScore{94.15}
      & \SecondScore{89.52}
      & \SecondScore{89.93}
      & \SecondScore{82.54} \\
      DIME-L
      & \BestScore{\textbf{94.27}}
      & \BestScore{\textbf{89.71}}
      & \BestScore{\textbf{90.12}}
      & \BestScore{\textbf{82.80}} \\
      \bottomrule
    \end{tabular*}

    \vspace{0.7mm}
    \parbox[t][1.55cm][t]{\linewidth}{%
      \footnotesize
      \textbf{(c) Face parsing.}
      Mean F1 and mIoU (\%) over foreground classes; CelebM denotes CelebAMask-HQ. In all panels, ``--'' marks results not reported by the original papers.
    }
  \end{minipage}
  \vspace{-2.0em}
\end{table*}

\noindent\textbf{Experimental protocols.} For landmark detection, we train and evaluate on the WFLW~\cite{wu2018look} dataset, which contains 98 landmarks and challenging variations in pose, expression, makeup, occlusion, and blur. All DIME variants use the heatmap-regression head and $448\times448$ inputs of FaRL-448-ft~\cite{zheng2022general}, the most comparable baseline. We report inter-ocular normalized mean error (NME), failure rate (FR), and area under the cumulative error distribution curve (AUC).

For head pose estimation, we use a common protocol: train on 300W-LP~\cite{zhu2019face} and evaluate on AFLW2000-3D~\cite{zhu2019face} and BIWI~\cite{fanelli2013random}. We adopt a linear six-dimensional head with an $\mathrm{SO}(3)$ geodesic loss~\cite{hempel20226d} and $224$px inputs. Performance is measured by the mean absolute angular error, in degrees, for yaw, pitch, and roll. Mean is the average of the three.

For face parsing, all the models are trained and evaluated separately on LaPa~\cite{liu2020new} and CelebAMask-HQ~\cite{lee2020maskgan}. LaPa contains 18,176/2,000/2,000 training/validation/test images and 11 semantic classes, whereas CelebAMask-HQ contains 24,183/2,993/2,824 images and 19 classes. We employ a shared multi-scale parsing head with $448\times448$ inputs and $512\times512$ predictions. Mean F1 and mIoU are macro-averaged over foreground classes. Within each task, DIME uses identical downstream architectures and optimization settings. We use EMA weights for evaluation without any test-time adaptation or model ensembling.

\noindent\textbf{Results on landmark detection.} In Table~\ref{tab:facial_geometry}(a), DIME-S is competitive with facial representation models and landmark-specific methods. NME falls from 3.93 (DIME-S) to 3.82 (DIME-B) and 3.74 (DIME-L), and AUC rises from 61.28 to 62.94 and 63.74; DIME-L improves the previous best NME from 3.86 to 3.74 and AUC from 62.40 to 63.74. DIME-B and DIME-L have an FR of 1.72\%, close to the 1.76\% of FaRL-448-ft and ORFormer. In the matched ablation, the full objective also lowers WFLW NME relative to single-image MAE on the same data (4.02 vs.\ 4.31, Table~\ref{tab:dime_ablation}).

\noindent\textbf{Results on head pose estimation.} In Table~\ref{tab:facial_geometry}(b), DIME-S lowers the mean error from 3.26$^\circ$ to 3.13$^\circ$ on AFLW2000-3D and from 3.66$^\circ$ to 3.45$^\circ$ on BIWI. On AFLW2000-3D, DIME-B is best with 3.01$^\circ$, 0.25$^\circ$ (7.7\% relative) below SRHP-Euler. On BIWI, DIME-L is best with 3.27$^\circ$, 0.39$^\circ$ (10.7\% relative) below MNN. DIME-B and DIME-L differ by at most 0.07$^\circ$ in mean error on either benchmark. The largest gain is in AFLW2000-3D pitch, from 4.25$^\circ$ for SRHP-Euler to 3.82$^\circ$.

\noindent\textbf{Results on face parsing.} Table~\ref{tab:facial_geometry}(c) evaluates dense facial semantics and part boundaries. DIME-S achieves 93.86\% mean F1 and 88.96\% mIoU on LaPa, and 89.61\% mean F1 and 81.96\% mIoU on CelebAMask-HQ. Both metrics improve consistently with model scale. DIME-L reaches 94.27\% mean F1 and 89.71\% mIoU on LaPa, exceeding the previous best results by 0.36 and 1.57 points, respectively, and obtains 90.12\% mean F1 and 82.80\% mIoU on CelebAMask-HQ, improving them by 0.49 and 1.25 points. The larger gains in mIoU are consistent with improved spatial overlap for compact or under-represented facial parts, suggesting that DIME preserves local boundaries and relative facial configuration beyond global appearance.

\noindent\textbf{Discussion.} DIME transfers consistently across sparse landmarks, 3D orientation, and dense semantic layout. This breadth suggests that DIME preserves geometrically meaningful facial variation, capturing how facial structure, head pose, and local parts vary across subjects and viewpoints.

\subsection{DIME for Face Generation}
\label{sec:exp_generation}

We investigate DIME's capacity to serve as a representation encoder for face generation~\cite{melnik2024face,jiang2025towards}. Representation-based generative models (e.g., latent image diffusion) rely on the encoder to define the latent space in which generation is performed. A useful facial representation should therefore preserve subtle cues that determine perceptual realism.

\begin{figure*}[t]
  \centering
  \setlength{\abovecaptionskip}{4pt}
  \includegraphics[width=1.0\textwidth]{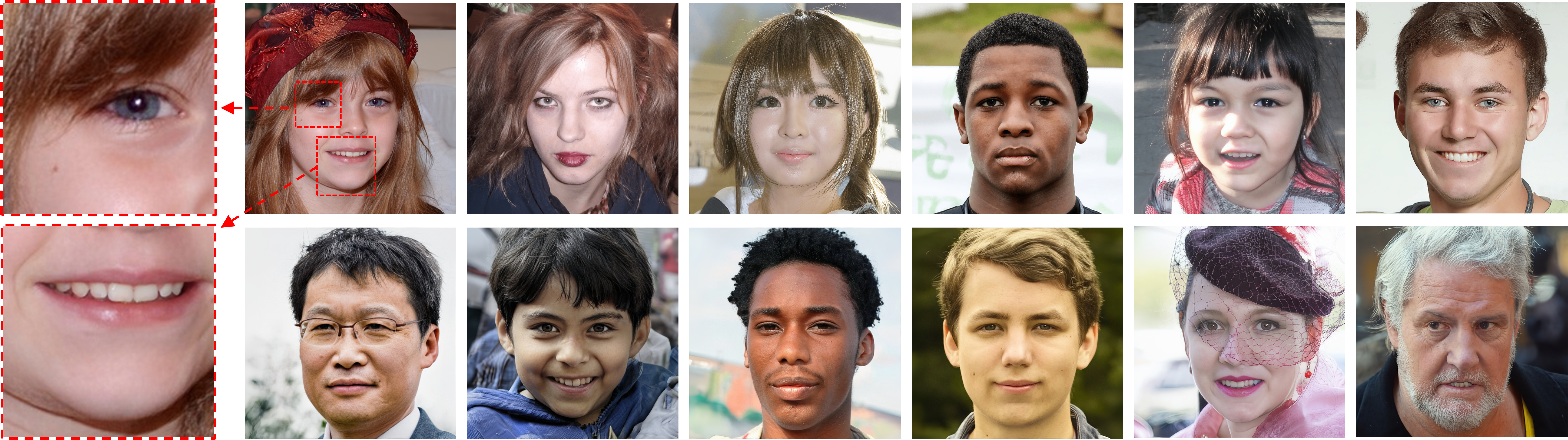}
\caption{Selected face images generated with the DIME-L encoder in the RAEv2 framework with 100K training steps.}
\vspace{-1.5em}
  \label{fig:face_generation_samples}
\end{figure*}

\noindent\textbf{Experimental setting.} We evaluate DIME on FFHQ~\cite{karras2019style} with two state-of-the-art latent diffusion frameworks: RAEv2~\cite{singh2026improved}, which constructs a generative latent space from a pretrained representation encoder, and REPA~\cite{yu2024representation}, which aligns the internal diffusion features with external representations. Both frameworks keep the encoder frozen.

We compare DIME with the generic encoders DINOv2~\cite{oquab2023dinov2}, DINOv3~\cite{simeoni2025dinov3}, MAE~\cite{he2022masked}, and MixMAE~\cite{liu2023mixmae} (public checkpoints), and with MAE-Face~\cite{ma2024facial}, which applies the standard MAE objective to face images.

\noindent\textbf{Generation quality.} At 100K steps, DIME-L has the lowest FID in all four settings of Table~\ref{tab:face_generation}: 6.12 and 6.59 with RAEv2 and 7.26 and 7.93 with REPA at $256$ and $512$, which are 51.0\%, 55.6\%, 46.1\%, and 50.9\% lower than DINOv3-L at the same 100K steps. At 50K steps, DIME-L (8.69, 9.41, 8.85, and 10.16) already has lower FID than every non-DIME encoder at 100K steps, although DINOv3-L trained for 400K steps remains lower in all four settings. At the same 100K-step budget, DIME-B also has lower FID than the larger generic encoders; under REPA, DINOv3-L at 400K steps (8.22 and 8.36) is still below DIME-B at 100K steps (8.47 and 9.25).

\begin{table}[!t]
  \centering
  \caption{Face-generation FID~\cite{heusel2017gans} (lower is better) on FFHQ at $256{\times}256$ and $512{\times}512$ under RAEv2~\cite{singh2026improved} and REPA~\cite{yu2024representation}. Gray arrows give the FID change relative to DINOv3-L at 100K steps. Best and second-best results are in \BestScore{vermilion} and \SecondScore{deep blue}, respectively. Steps: training steps.}
  \label{tab:face_generation}
  \setlength{\tabcolsep}{5.0pt}
  \renewcommand{\arraystretch}{1.0}
  \resizebox{\columnwidth}{!}{%
  \begin{tabular}{@{}l c r r r r@{}}
    \toprule
    & & \multicolumn{2}{c}{RAEv2 \cite{singh2026improved}} & \multicolumn{2}{c}{REPA \cite{yu2024representation}} \\
    \cmidrule(lr){3-4}\cmidrule(l){5-6}
    Rep. encoder & Steps
      & \multicolumn{1}{c}{256/FID} & \multicolumn{1}{c}{512/FID}
      & \multicolumn{1}{c}{256/FID} & \multicolumn{1}{c}{512/FID} \\
    \midrule
    DINOv3-L \cite{simeoni2025dinov3}     & 100K & \multicolumn{1}{c}{12.48} & \multicolumn{1}{c}{14.83}
                       & \multicolumn{1}{c}{13.46} & \multicolumn{1}{c}{16.15} \\
    DINOv3-L \cite{simeoni2025dinov3}    & 200K &  9.83\fiddown{2.65} & 11.53\fiddown{3.30}
                       & 10.61\fiddown{2.85} & 11.80\fiddown{4.35} \\
    DINOv3-L \cite{simeoni2025dinov3}    & 400K &  8.53\fiddown{3.95} &  8.94\fiddown{5.89}
                       & \SecondScore{8.22}\fiddown{5.24} & \SecondScore{8.36}\fiddown{7.79} \\
    \midrule
    DINOv2-L \cite{oquab2023dinov2}     & 100K & 14.83\fidup{2.35} & 16.79\fidup{1.96} & 15.31\fidup{1.85} & 18.82\fidup{2.67} \\
    MAE-L \cite{he2022masked}        & 100K & 16.72\fidup{4.24} & 18.22\fidup{3.39} & 18.34\fidup{4.88} & 21.31\fidup{5.16} \\
    MixMAE-L \cite{liu2023mixmae}     & 100K & 17.44\fidup{4.96} & 19.65\fidup{4.82} & 17.15\fidup{3.69} & 20.83\fidup{4.68} \\
    MAE-Face \cite{ma2024facial}     & 100K & 11.03\fiddown{1.45} & 16.50\fidup{1.67} & 13.67\fidup{0.21} & 18.67\fidup{2.52} \\
    \midrule
    DIME-B       &  50K &  9.52\fiddown{2.96} &  9.98\fiddown{4.85} & 10.92\fiddown{2.54} & 11.03\fiddown{5.12} \\
    DIME-B       & 100K & \SecondScore{7.41}\fiddown{5.07} &
                            \SecondScore{8.06}\fiddown{6.77} &
                            8.47\fiddown{4.99} &
                            9.25\fiddown{6.90} \\
    DIME-L       &  50K &  8.69\fiddown{3.79} &  9.41\fiddown{5.42} &  8.85\fiddown{4.61} & 10.16\fiddown{5.99} \\
    DIME-L       & 100K & \BestScore{\textbf{6.12}}\fiddown{6.36} &
                            \BestScore{\textbf{6.59}}\fiddown{8.24} &
                            \BestScore{\textbf{7.26}}\fiddown{6.20} &
                            \BestScore{\textbf{7.93}}\fiddown{8.22} \\
    \bottomrule
  \end{tabular}%
  }
  \vspace{-2.0em}
\end{table}

\begin{figure}[t]
  \centering
  \setlength{\abovecaptionskip}{0pt}
  \includegraphics[width=1.0\linewidth]{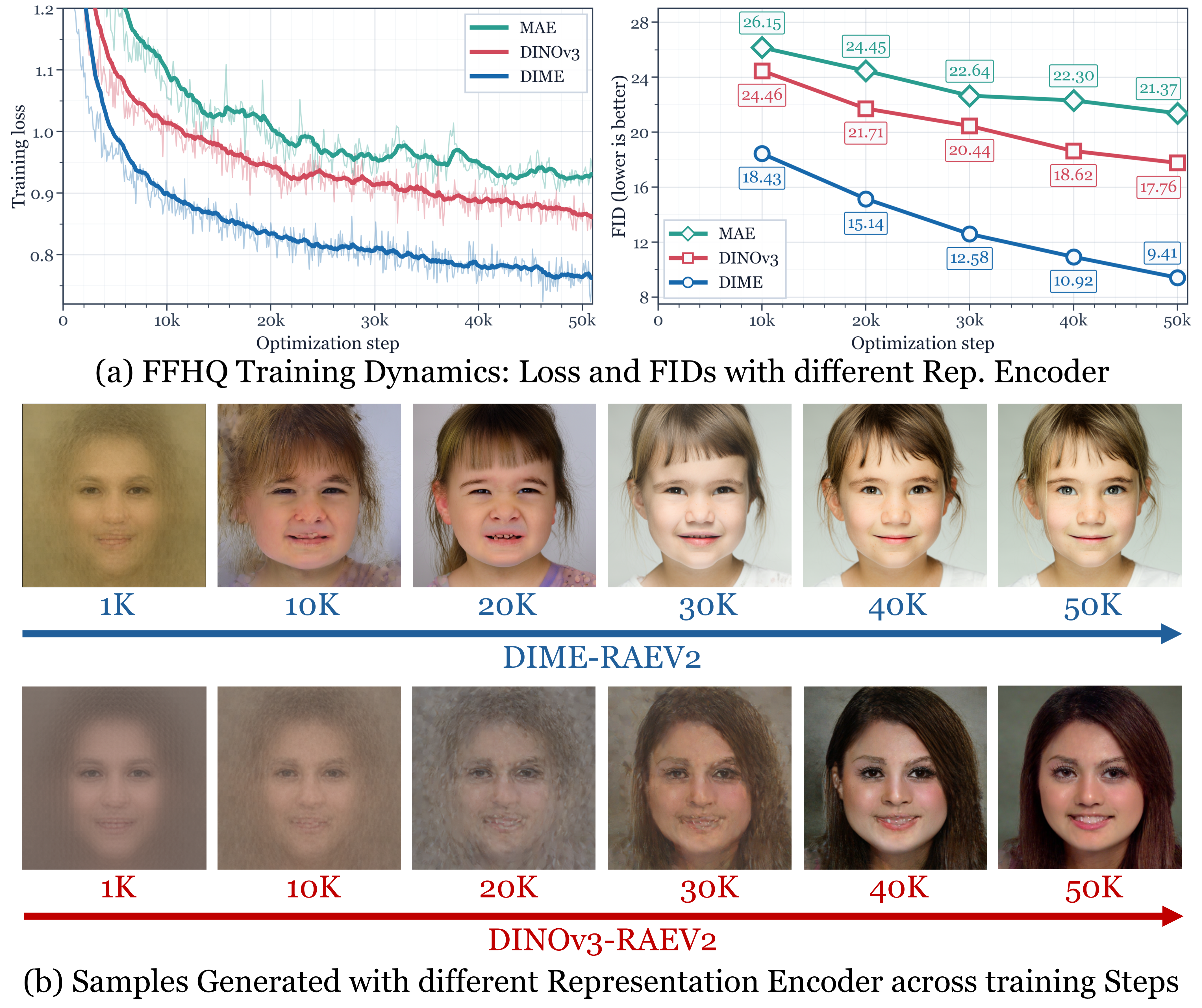}
\caption{\textbf{Face-generation dynamics on FFHQ.} (a) Training loss and FID of RAEv2 using MAE, DINOv3, and DIME encoders. DIME consistently converges faster and achieves lower FID. (b) Generated samples from 1K to 50K training steps using DIME (top) and DINOv3 (bottom). DIME produces coherent facial structure earlier and resolves fine details sooner, consistent with its faster FID reduction.}
    \label{fig:face_generation_dynamics}
  \vspace{-0.5em}
\end{figure}

\noindent\textbf{Convergence and resolution.} Fig.~\ref{fig:face_generation_dynamics}(a) compares RAEv2 with MAE-L, DINOv3-L, and DIME-L encoders at $512\times512$ under the same training setting. DIME-L has lower training loss and FID throughout the first 50K steps. At 10K steps its FID is 18.43, compared with 24.46 for DINOv3-L and 26.15 for MAE-L; at 50K steps the values are 9.41, 17.76, and 21.37, so at 50K steps DIME-L is already below DINOv3-L at 200K steps (11.53; Table~\ref{tab:face_generation}). From 50K to 100K steps, the four FID scores improve by 16.1--22.4\% for DIME-B and 18.0--30.0\% for DIME-L. For DIME-L at 100K steps, raising the output resolution from 256 to 512 increases FID by 0.47 with RAEv2 and 0.67 with REPA, compared with 1.50--5.47 for the non-DIME encoders at 100K steps; DINOv3-L, trained four times longer (400K steps), shows a smaller increase (0.41 and 0.14).

\noindent\textbf{Generated samples.}
Fig.~\ref{fig:face_generation_samples} presents samples generated by RAEv2 with the DIME-L encoder. Fig.~\ref{fig:face_generation_dynamics}(b) compares generated samples during training. Both DIME and DINOv3 produce coarse facial layouts at 1K steps, but DIME resolves recognizable eyes, mouth, and hair by 10K--20K steps, while the DINOv3 samples remain blurred or visibly noisy. From 30K to 50K steps, DIME further refines facial contours and local details, while DINOv3 produces coarser facial images. This qualitative comparison is consistent with the lower FID of DIME in panel~(a), supporting faster acquisition of both global facial structure and fine-grained appearance.

\begin{table*}[t]
  \centering
  \setlength{\abovecaptionskip}{2pt}
  \caption{\textbf{Controlled ablation study of DIME.}
  Panels (a)--(c) examine the pretraining unit, the interaction between identity matching and EDDS, and source isolation.
  Panels (d)--(f) evaluate complementary supervision, facial-change information retained in frozen features, and sensitivity to the source-mixing ratio. All variants: ViT-S/16, 100 epochs, 30\% of Face-20M.}
  \label{tab:dime_ablation}

  \begingroup
  \scriptsize
  \setlength{\aboverulesep}{0.35ex}
  \setlength{\belowrulesep}{0.35ex}
  \definecolor{dimeDefault}{gray}{0.92}
  \def\ablpaneltitle#1{%
    \parbox[c][1.35\baselineskip][c]{\linewidth}{%
      \centering\bfseries\strut #1\strut}\par
    \vspace{-0.10\baselineskip}%
  }

  \begin{minipage}[t]{0.323\textwidth}
    \vspace{0pt}
    \centering
    \ablpaneltitle{(a) Pretraining unit under RGB recovery}
    {\renewcommand{\arraystretch}{1.00}
    \setlength{\tabcolsep}{1.0pt}
    \begin{tabular*}{\linewidth}{@{\extracolsep{\fill}}lcccc@{}}
      \toprule
      & \multicolumn{4}{c}{Transfer} \\
      \cmidrule(l){2-5}
      Unit & F1$\uparrow$ & ICC$\uparrow$ & NME$\downarrow$ & HTER$\downarrow$ \\
      \midrule
      Single (MAE~\cite{he2022masked}) & 61.9 & .619 & 4.31 & 11.32 \\
      Same image & 62.3 & .623 & 4.29 & 10.91 \\
      Random ID & 62.8 & \textbf{.630} & 4.23 & 10.38 \\
      \rowcolor{dimeDefault}
      Same ID & \textbf{62.9} & \textbf{.630} & \textbf{4.22} & \textbf{10.32} \\
      \midrule
      Same--random & $+0.1$ & $+.001$ & $-.01$ & $-0.06$ \\
      \bottomrule
    \end{tabular*}}
  \end{minipage}\hfill%
  \begin{minipage}[t]{0.323\textwidth}
    \vspace{0pt}
    \centering
    \ablpaneltitle{(b) Identity control $\times$ EDDS}
    {\renewcommand{\arraystretch}{1.00}
    \setlength{\tabcolsep}{0.7pt}
    \begin{tabular*}{\linewidth}{@{\extracolsep{\fill}}lccccc@{}}
      \toprule
      & & \multicolumn{4}{c}{Transfer} \\
      \cmidrule(l){3-6}
      Pair & E & F1$\uparrow$ & ICC$\uparrow$ & NME$\downarrow$ & HTER$\downarrow$ \\
      \midrule
      Random ID & $-$ & 62.8 & .630 & 4.23 & 10.38 \\
      Random ID & \checkmark & 63.1 & .630 & 4.24 & 10.57 \\
      Same ID & $-$ & 62.9 & .630 & 4.22 & 10.32 \\
      \rowcolor{dimeDefault}
      Same ID & \checkmark & \textbf{65.1} & \textbf{.651} & \textbf{4.07} & \textbf{8.64} \\
      \midrule
      \multicolumn{2}{l}{Interaction} & $+1.9$ & $+.020$ & $-.16$ & $-1.87$ \\
      \bottomrule
    \end{tabular*}}
  \end{minipage}\hfill%
  \begin{minipage}[t]{0.323\textwidth}
    \vspace{0pt}
    \centering
    \ablpaneltitle{(c) Pairing $\times$ source isolation}
    {\renewcommand{\arraystretch}{1.00}
    \setlength{\tabcolsep}{0.6pt}
    \begin{tabular*}{\linewidth}{@{\extracolsep{\fill}}llcccc@{}}
      \toprule
      & & \multicolumn{4}{c}{Transfer} \\
      \cmidrule(l){3-6}
      Pair & Enc. isol. & F1$\uparrow$ & ICC$\uparrow$ & NME$\downarrow$ & HTER$\downarrow$ \\
      \midrule
      Random & None & 62.5 & .620 & 4.32 & 11.15 \\
      Random & All & 63.1 & .630 & 4.24 & 10.57 \\
      Same & None & 63.4 & .634 & 4.20 & 10.06 \\
      \rowcolor{dimeDefault}
      Same & All & \textbf{65.1} & \textbf{.651} & \textbf{4.07} & \textbf{8.64} \\
      \midrule
      \multicolumn{2}{l}{Interaction} & $+1.1$ & $+.007$ & $-.05$ & $-0.84$ \\
      \bottomrule
    \end{tabular*}}
  \end{minipage}

  \par\vspace{2.5mm}

  \begin{minipage}[t][37mm][t]{0.323\textwidth}
    \vspace{0pt}
    \centering
    \ablpaneltitle{(d) Objective--task correspondence}
    {\renewcommand{\arraystretch}{1.03}
    \setlength{\tabcolsep}{1.3pt}
    \begin{tabular*}{\linewidth}{@{\extracolsep{\fill}}cccccc@{}}
      \toprule
      & & \multicolumn{4}{c}{Transfer} \\
      \cmidrule(l){3-6}
      G & E & F1$\uparrow$ & ICC$\uparrow$ & NME$\downarrow$ & HTER$\downarrow$ \\
      \midrule
      $-$ & $-$ & 62.9 & .630 & 4.22 & 10.32 \\
      \checkmark & $-$ & 63.8 & .653 & 4.04 & 9.83 \\
      $-$ & \checkmark & 65.1 & .651 & 4.07 & 8.64 \\
      \rowcolor{dimeDefault}
      \checkmark & \checkmark & \textbf{65.8} & \textbf{.663} & \textbf{4.02} & \textbf{8.11} \\
      \midrule
      \multicolumn{2}{l}{Avg. $\Delta_{\rm G}$} & $+0.8$ & $+.018$ & $-.12$ & $-0.51$ \\
      \multicolumn{2}{l}{Avg. $\Delta_{\rm E}$} & $+2.1$ & $+.016$ & $-.09$ & $-1.70$ \\
      \bottomrule
    \end{tabular*}}
    \vfill
  \end{minipage}\hfill%
  \begin{minipage}[t][37mm][t]{0.323\textwidth}
    \vspace{0pt}
    \centering
    \ablpaneltitle{(e) Frozen facial-change probe}
    {\renewcommand{\arraystretch}{1.03}
    \setlength{\tabcolsep}{2.0pt}
    \begin{tabular*}{\linewidth}{@{\extracolsep{\fill}}lccc@{}}
      \toprule
      & & \multicolumn{2}{c}{Change probe} \\
      \cmidrule(l){3-4}
      Pretraining & Pair & F1$\uparrow$ & ICC$\uparrow$ \\
      \midrule
      MAE~\cite{he2022masked} & $-$ & 43.4 & .298 \\
      $R$ & Same & 45.5 & .326 \\
      $R+$G & Same & 47.6 & .357 \\
      $R+$E & Random & 46.2 & .337 \\
      $R+$E & Same & 52.0 & .405 \\
      \rowcolor{dimeDefault}
      Full ($R+$G$+$E) & Same & \textbf{53.4} & \textbf{.424} \\
      \bottomrule
    \end{tabular*}}
    \vfill
  \end{minipage}\hfill%
  \begin{minipage}[t][37mm][t]{0.323\textwidth}
    \vspace{0pt}
    \centering
    \ablpaneltitle{(f) Source-mixing ratio}
    {\renewcommand{\arraystretch}{1.03}
    \setlength{\tabcolsep}{1.3pt}
    \begin{tabular*}{\linewidth}{@{\extracolsep{\fill}}ccccc@{}}
      \toprule
      & \multicolumn{4}{c}{Transfer} \\
      \cmidrule(l){2-5}
      Ratio $r$ & F1$\uparrow$ & ICC$\uparrow$ & NME$\downarrow$ & HTER$\downarrow$ \\
      \midrule
      $.10$ & 63.1 & .634 & 4.20 & 10.25 \\
      $.20$ & 64.0 & .643 & 4.13 & 9.57 \\
      $.30$ & 64.9 & .653 & 4.07 & 8.78 \\
      $.40$ & 65.7 & \textbf{.664} & 4.03 & 8.20 \\
      \rowcolor{dimeDefault}
      $.50^{\dagger}$ & \textbf{65.8} & .663 & \textbf{4.02} & \textbf{8.11} \\
      \midrule
      $\Delta_{.50-.10}$ & $+2.7$ & $+.029$ & $-.18$ & $-2.14$ \\
      \bottomrule
    \end{tabular*}}
    \vfill
  \end{minipage}

  \par\vspace{-1.5em}
  \begin{minipage}{0.985\textwidth}
    \scriptsize
    \emph{Notes.}
    $R$ denotes reciprocal RGB recovery with all-block source isolation;
    G and E denote gradient-orientation supervision and EDDS.
    Derived rows are computed from unrounded means and are not additional models.
    Interaction is the same-identity treatment effect minus the random-identity treatment effect:
    enabling E in (b) and enabling source isolation in (c).
    In (c), ``None'' removes the encoder source-attention mask, whereas ``All'' applies it in every block.
    In (e), ``Pair'' denotes pretraining pairing; all change probes use same-subject pairs.
    In (f), $r$ is the counterpart-patch fraction; reciprocal exchange makes $r$ and $1-r$ equivalent.
    Gray rows mark the reference configurations within each panel, and $\dagger$ marks the default ratio.
    Boldface identifies the best displayed results among model rows.
  \end{minipage}
  \endgroup
  \vspace{-1.0em}
\end{table*}

\noindent\textbf{Discussion.} DIME emphasizes how facial observations vary, encouraging the representation to capture fine-grained factors that change across observations. Such representations provide a more informative space for face generation, leading to faster convergence and better generation quality.

\subsection{Ablation Studies and Analysis}
\label{sec:exp_ablation}

\noindent\textbf{Controlled protocol.}
Table~\ref{tab:dime_ablation} shows controlled ablation studies of DIME. To reduce compute, we choose ViT-S/16 as the encoder backbone and pretrain it for 100 epochs on a 30\% subset of Face-20M, holding the image pool, training settings, and decoder configuration fixed. Paired variants use a 50\% source partition except in panel~(f). Random-identity controls re-pair the same images across different identities, preserving the image pool while changing the relation between observations. We denote reciprocal RGB recovery with all-block source isolation by $R$, gradient-orientation supervision by G, and EDDS by E.

Panels~(a)--(c) examine the learning unit and the conditions under which pairing becomes useful. Panel~(a) uses RGB supervision only, comparing single-image MAE with paired recovery from two augmentations of one image, distinct images of different identities, or distinct images of the same identity. Panel~(b) crosses random- versus same-identity pairing with EDDS disabled or enabled, testing whether differential supervision specifically benefits from a shared identity. Panel~(c) keeps EDDS enabled and compares unrestricted encoder attention (``None'') with source isolation in every block (``All'') for both pairing strategies. G is disabled throughout panels~(a)--(c) to isolate these factors.

Panels~(d)--(f) examine objective contributions, retained facial-change information, and mixing sensitivity. Panel~(d) fixes same-identity pairing and source isolation, adding G and E separately and jointly to $R$. Panel~(e) freezes the pretrained encoders and trains linear probes on feature differences to test whether facial changes remain accessible without encoder adaptation. Panel~(f) retains same-identity pairing, source isolation, and the full objective, varying only the counterpart-patch fraction $r$ of each grid from $.10$ to $.50$.

For downstream evaluations, we follow the BP4D~\cite{zhang2014bp4d} and DISFA~\cite{mavadati2013disfa} folds used in the main experiments, with three fine-tuning seeds for WFLW~\cite{wu2018look} and each FAS target. Panels~(a)--(d) and~(f) report BP4D mean AU F1, DISFA mean ICC(3,1), WFLW inter-ocular NME, and macro HTER over four cross-domain transfers; panel~(e) evaluates facial-change prediction with frozen encoders. All results use the reduced data and pretraining protocol and are intended for comparisons within the ablation study only.

\noindent\textbf{Identity matching needs EDDS.}
In panel~(a), paired RGB recovery slightly improves over single-image MAE, but replacing random- with same-identity pairs yields only marginal gains. Identity matching alone therefore provides little benefit when supervision decomposes into separate losses. With EDDS enabled in panel~(b), same-identity pairing outperforms random pairing by $2.0$ F1, $.021$ ICC, $.17$ NME, and $1.93$ HTER. The interaction effects measure how much more EDDS benefits same-identity pairs than random pairs and favor identity matching across all four metrics. These results support their joint role: matching establishes a shared appearance reference, while differential supervision makes the remaining variation relevant to optimization.

\noindent\textbf{Why source isolation matters.}
Panel~(c) tests whether paired supervision remains effective when the encoder can attend across observations. All-block source isolation improves every metric, with larger gains for same-identity pairs: $+1.7$ F1 and $-1.42$ HTER, compared with $+0.6$ and $-0.58$ for random pairs. ICC and NME follow the same pattern. The larger benefit under identity matching is consistent with shared appearance providing greater opportunity for cross-observation shortcuts. Source isolation requires each reconstruction to depend on its own visible evidence, strengthening the representation retained for single-image transfer.

\noindent\textbf{Complementary objective contributions.}
Panel~(d) evaluates the complete $2\!\times\!2$ combination of G and E. Each marginal effect averages one objective's contribution with the other disabled and enabled. Relative to E, G contributes more to AU intensity estimation and landmark localization, increasing ICC by $.018$ and reducing NME by $.12$ on average. E contributes more to AU detection and cross-domain FAS, increasing F1 by $2.1$ and reducing HTER by $1.70$. Combining both achieves the best results on all four metrics. Although the additional NME reduction over G alone is small, the results support complementary roles for local-geometry recovery and cross-image differential supervision.

\noindent\textbf{Frozen facial-change probes.}
Panel~(e) tests the accessibility of facial variation from the frozen encoder. The ``Pair'' column denotes pretraining pairing; all probes use same-subject frame pairs. We sample a fixed number of pairs across predefined temporal offsets, assign the earlier frame to $a$ and the later frame to $b$, and balance decrease, no-change, and increase transitions. Each image is encoded independently, and the probe receives only $\Delta\mathbf{z}=\mathbf{z}^{b}-\mathbf{z}^{a}$. BP4D per-AU linear classifiers predict decrease, no change, or increase in AU occurrence and are evaluated by F1. DISFA per-AU linear regressors predict signed intensity differences and are evaluated by ICC(3,1). Under $R+$E, same-identity pretraining improves change-prediction F1 from 46.2 to 52.0 and ICC from $.337$ to $.405$ over random pairing. The full model reaches 53.4 and $.424$, compared with 43.4 and $.298$ for MAE. These gains show that facial-change information is retained in the encoder and remains accessible without downstream feature adaptation.

\noindent\textbf{Sensitivity to source mixing.}
Panel~(f) evaluates the setting of the counterpart-patch fraction $r$ under the full objective. Transfer improves from $.10$ to $.40$, with only small changes between $.40$ and $.50$. Across the tested range, F1 increases by $2.7$ and HTER decreases by $2.14$. The $.40$ setting gives the highest ICC, while $.50$ gives the best F1, NME, and HTER. Performance is therefore stable near balanced mixing; we use $r=.50$ for source symmetry by default.

\noindent\textbf{Discussion.}
The ablations connect DIME's construction to its learned representations: identity matching provides little benefit under independent recovery, becomes effective through EDDS, and benefits further from source isolation. Frozen probes confirm that the encoder preserves more accessible facial-change information. These findings explain how DIME makes within-identity variation useful for pretraining, complementing the data-scaling results in Fig.~\ref{fig:1}(d).

\section{Limitations and Future Work}
\label{sec:discussion}

\noindent\textbf{Can scale alone solve the problem?}
Scaling laws~\cite{kaplan2020scaling,hoffmann2022training} assume an objective that keeps extracting new signal from more data. On faces, MAE-B gains only about 0.5 F1 from 6M to 20M images, whereas DIME-B gains 5.9 (Fig.~\ref{fig:1}(d)); scale thus amplifies the objective rather than replacing it. DIME shows no plateau up to 20M but will likely saturate once the within-identity variation is exhausted, as its effective scale is variation diversity rather than image count~\cite{sorscher2022beyond}. Model scaling also shows diminishing returns (BP4D F1: $+2.2$ from DIME-S to DIME-B, $+1.2$ from DIME-B to DIME-L); a joint study of variation diversity, model size, and training length~\cite{hoffmann2022training,xie2023data} is left for future work.

\noindent\textbf{Where and why does DIME fail?}
We identify three sources. (i) \emph{Benchmark ceilings}: on AUs above 85\% F1 (e.g., AU10 and AU12), DIME-L stays within 1.4 points of MAE-Face and all the compared methods cluster closely. (ii) \emph{Data coverage}: all facial-pretrained models, including DIME, underperform generic encoders on DISFA AU4 (Table~\ref{tab:au_disfa}). Moreover, Face-20M's portraits rarely contain large pitch changes, consistent with residual weaknesses on BIWI pitch; training on more same-distribution images is unlikely to help. (iii) \emph{Method design}: for simplicity, EDDS supervises every within-identity change that the context predicts, including pose, illumination, and misalignment (Sec.~\ref{sec:dime_formal}), rather than task-relevant change alone. Future gains thus lie in more diverse within-identity variation (e.g., video), in pose-matched pairs, and in objectives that isolate task-relevant change.

\noindent\textbf{Future work.}
Fine-grained facial analysis remains limited by supervision: dense AU annotation requires trained coders, while privacy laws increasingly restrict identity-labeled face datasets. DIME uses identity only to establish that the paired facial images depict the same person. Its training pairs can therefore come from videos collected with consent, multi-view captures, or generated faces of the same identity. Our generation results suggest a training loop in which a DIME encoder improves a generator that then supplies controlled pairs, such as faces with matching poses that differ in a single AU. DIME could also serve as a facial encoder for multimodal language models, helping them distinguish AU-level changes in clinical and affective applications.

\section{Conclusion}

We presented DIME, a masked autoencoder that learns facial representations from identity-matched image pairs. DIME reconstructs two images of the same person, aligns the direction of their reconstructed difference with the true difference through EDDS, and uses source isolation so that each reconstruction depends only on its own image. Identity metadata is used only to form pairs; the training targets are pixels, image gradients, and pixel differences, and no facial-state, AU, or dense annotation is used. On Face-20M, DIME-B continues to improve BP4D AU detection up to 20M pretraining images, while MAE-B trained on the same subsets plateaus near 6M. In matched ablations, DIME improves over MAE on four tasks, and identity matching helps mainly in combination with EDDS and source isolation. We show that the pretrained DIME encoder transfers effectively to diverse facial analysis and generation tasks, including AU detection and intensity estimation, cross-domain anti-spoofing, facial geometry estimation, and face generation.

\bibliographystyle{IEEEtran}
\bibliography{reference}

\begin{IEEEbiography}[{\includegraphics[width=1in,height=1.25in,clip,keepaspectratio]{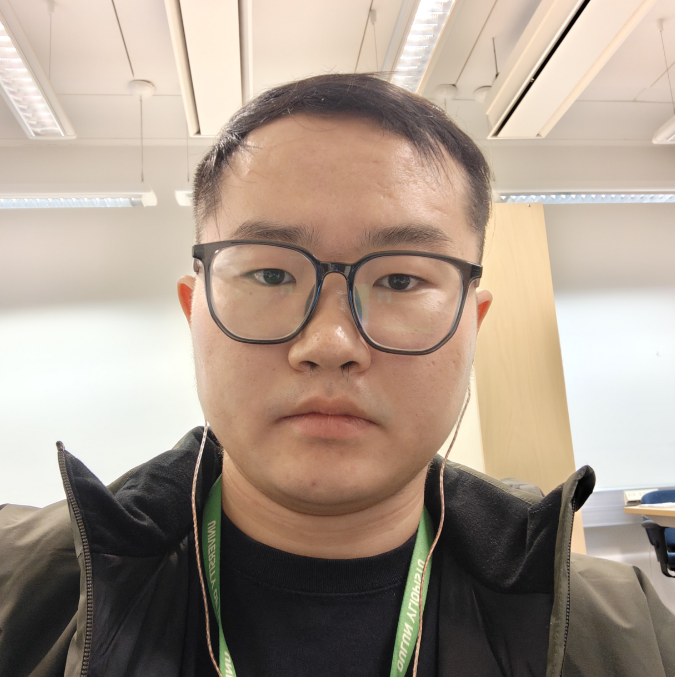}}]{Hao Yu} is currently a Ph.D. student at the Center for Machine Vision and Signal Analysis, University of Oulu, Finland. His research interests focus on visual representation learning, facial attribute analysis and visual generative models.
\end{IEEEbiography}

\begin{IEEEbiography}[{\includegraphics[width=1in,height=1.25in,clip,keepaspectratio]{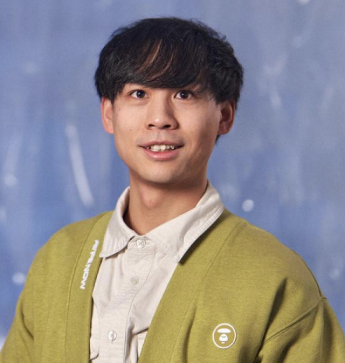}}]{Haoyu Chen}
received his Ph.D. degree in Computer Science and Engineering from the University of Oulu, Finland, in 2022. He is currently a Tenure Track Assistant Professor with CMVS and an Academy Research Fellow at the University of Oulu, Finland. He has actively chaired workshops and challenges at various conferences, including HHAI 2024, IJCAI 2023, and IJCAI 2024. His work on gesture recognition has received several awards, including the second prize of the IEEE Finland Jt. Chapter SP/CAS Best Paper Award, the 2nd place in the Action Recognition Track of the ECCV 2020 VIPriors Challenges, etc. His research interests include human behavior analysis, hybrid intelligence, and emotion AI.
\end{IEEEbiography}

\begin{IEEEbiography}[{\includegraphics[width=1in,height=1.25in,clip,keepaspectratio]{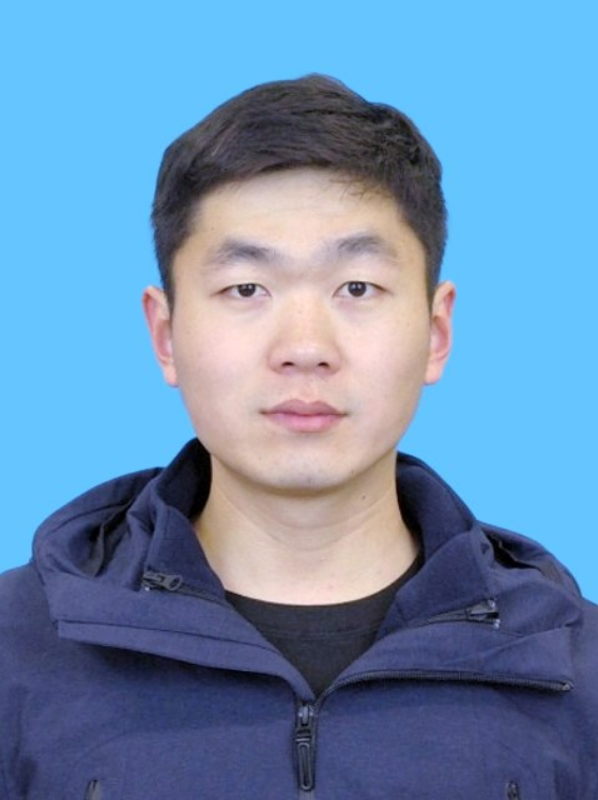}}]{Hui Wei}
received the Ph.D. degree in computer science from Wuhan University, China. He is currently a postdoctoral researcher with the Center for Machine Vision and Signal Analysis, University of Oulu, Finland. He has authored or coauthored papers in mainstream conferences and journals, including CVPR, NeurIPS, AAAI, ACM Multimedia, and the IEEE TPAMI. His research interests include trustworthy AI, privacy protection, and machine learning.
\end{IEEEbiography}

\begin{IEEEbiography}[{\includegraphics[width=1in,height=1.25in,clip,keepaspectratio]{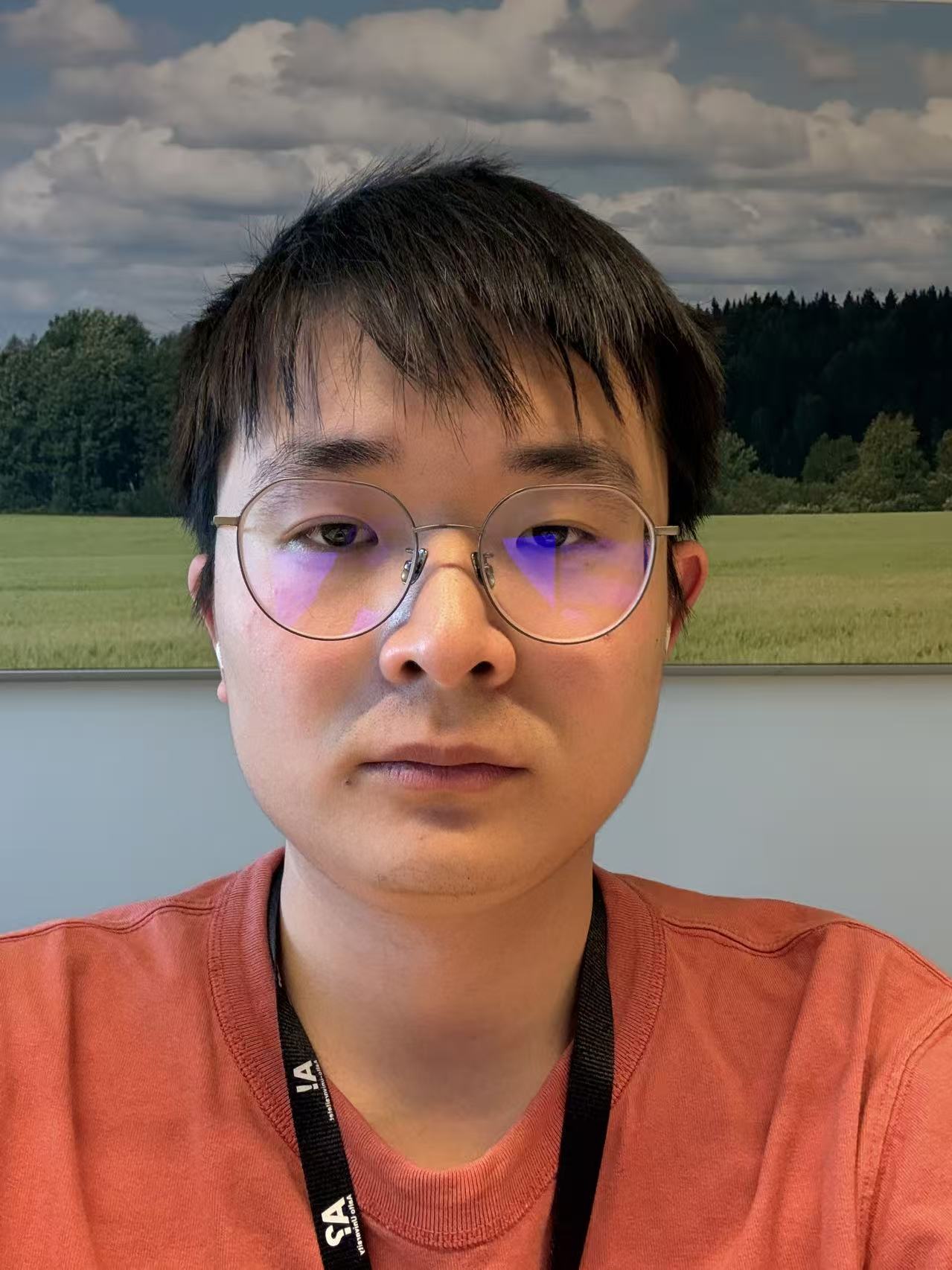}}]{Yan Jiang}
received the B.E. and M.S. degrees from Nanjing University of Information Science and Technology in 2022 and 2025, respectively. He is currently pursuing his Ph.D. degree in computer science at the University of Oulu, Finland. His research interests include human-centric vision and visual generative models.
\end{IEEEbiography}

\begin{IEEEbiography}[{\includegraphics[width=1in,height=1.25in,clip,keepaspectratio]{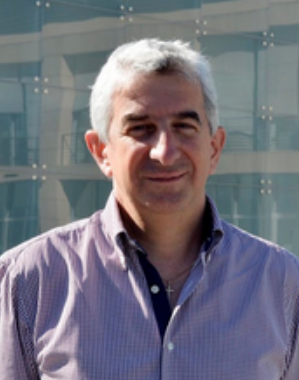}}]{Nicu Sebe} (Senior Member, IEEE) is a professor with the University of Trento, Italy, leading the research in the areas of multimedia information retrieval and human behavior understanding in Multimedia and Human Understanding Group (MHUG). He was the General Co-Chair of ACM Multimedia 2013, and the Program Chair of ACM Multimedia 2007 and 2011, ECCV 2016, ICCV 2017 and ICPR 2020. He is a fellow of the International Association for Pattern Recognition.
\end{IEEEbiography}

\begin{IEEEbiography}[{\includegraphics[width=1in,height=1.25in,clip,keepaspectratio]{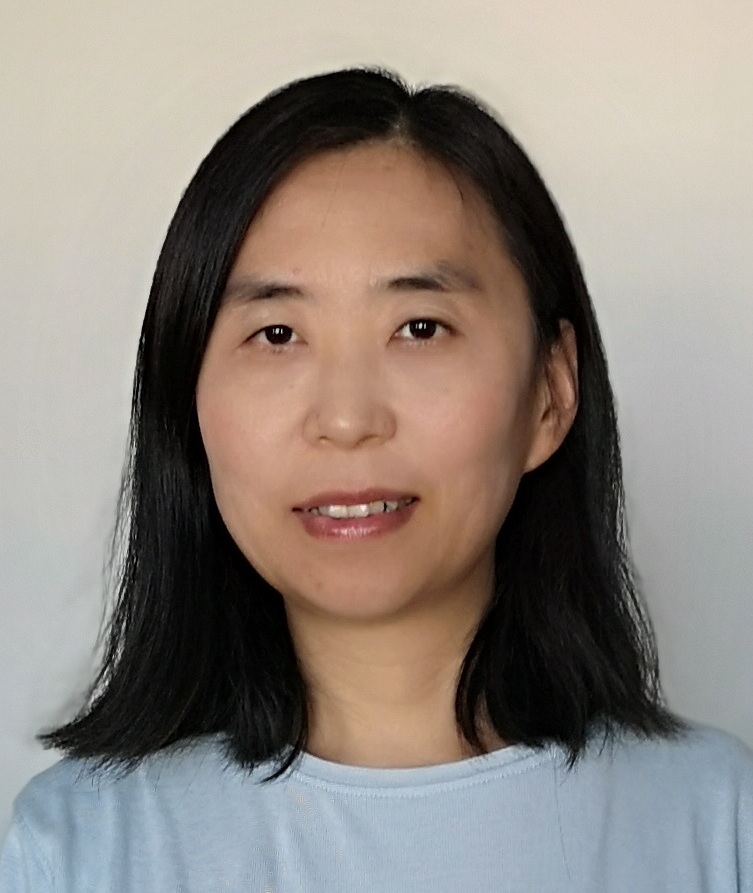}}]{Guoying Zhao} (IEEE Fellow 2022) received the Ph.D. degree in computer science from the Chinese Academy of Sciences, Beijing, China, in 2005. She is currently an Academy Professor and full Professor (tenured in 2017) with the University of Oulu, and a PI with ELLIS Institute Finland. She is a member of Academia Europaea, a member of the Finnish Academy of Sciences and Letters, Fellow of IEEE, IAPR, ELLIS and AAIA.
Her current research interests include image and video representation, facial-expression and micro-expression recognition, emotional gesture analysis, affective computing, and biometrics.
\end{IEEEbiography}

\end{document}